\documentclass[10pt,twocolumn,letterpaper]{article}
\usepackage[pagenumbers]{nomstyle}
\usepackage{amsmath,amssymb,booktabs,graphicx,xcolor,multirow}
\usepackage{algorithm}
\usepackage[noend]{algpseudocode}

\definecolor{linkblue}{rgb}{0.21,0.49,0.74}
\usepackage[pagebackref,breaklinks,colorlinks,allcolors=linkblue]{hyperref}

\newcommand{\E}{\mathbb{E}}
\newcommand{\Var}{\operatorname{Var}}

\title{The Cost of Omission: How Missing Class Names Degrade Open-Vocabulary Detection}
\author{\begin{tabular}{ccc}
Chen Zhang & Yifan Wang & Donglin Jia \\
Mengkun Guo & Jirong Ding & Xiaoyang Hao
\end{tabular}}
\hypersetup{pdftitle={The Cost of Omission: How Missing Class Names Degrade Open-Vocabulary Detection},
pdfauthor={Chen Zhang, Yifan Wang, Donglin Jia, Mengkun Guo, Jirong Ding, Xiaoyang Hao}}
\makeatletter
\def\nomsection{\@startsection{section}{1}{\z@}{-10pt plus -2pt minus -2pt}{7pt}{\large\bf\raggedright\hyphenpenalty=10000\exhyphenpenalty=10000}}
\def\nomsubsection{\@startsection{subsection}{2}{\z@}{-8pt plus -2pt minus -2pt}{5pt}{\elvbf\raggedright\hyphenpenalty=10000\exhyphenpenalty=10000}}
\def\nomsubsubsection{\@startsection{subsubsection}{3}{\z@}{-6pt plus -2pt minus -2pt}{3pt}{\tenbf\raggedright\hyphenpenalty=10000\exhyphenpenalty=10000}}
\makeatother
\begin{document}

\maketitle
\begin{abstract}
Open-vocabulary detectors are typically evaluated with a full class list, although users often ask for only a few classes. On COCO, shortening the list to eight classes can cost a text-fusion detector up to 16.2 AP50 points and can reverse the ordering of detectors. We propose NoM, a training-free method that adds scene names proposed from the image, filters out names closely related to the query, and restricts predictions to the queried classes. Controlled measurements motivate these steps. Adding names of objects present in the image improves recognition even when the detector cannot report those names, while the same number of absent names helps little. With the classifier fixed, removing an object's own name changes its visual representation and can cause it to be reported as a queried class. A first-order analysis predicts the resulting score change from the direction of the feature update more accurately than from the attention weight on the name alone. A name closely related to a queried class can even lower the score of a true instance of that class, which is why NoM filters such names. With eight queried classes and the same output restriction, NoM outperforms attention sinks and recovers 79\% and 66\% of the lost AP50 on Grounding DINO-L and LLMDet-L, respectively. In \mbox{Qwen2.5-VL}, a longer list can help when forming the image representation yet hurt when generating the answer. Code will be made publicly available.

\end{abstract}

\section{Introduction}
\begin{figure*}[t]
\centering\includegraphics[width=\textwidth,trim=0bp 1bp 0bp 1bp]{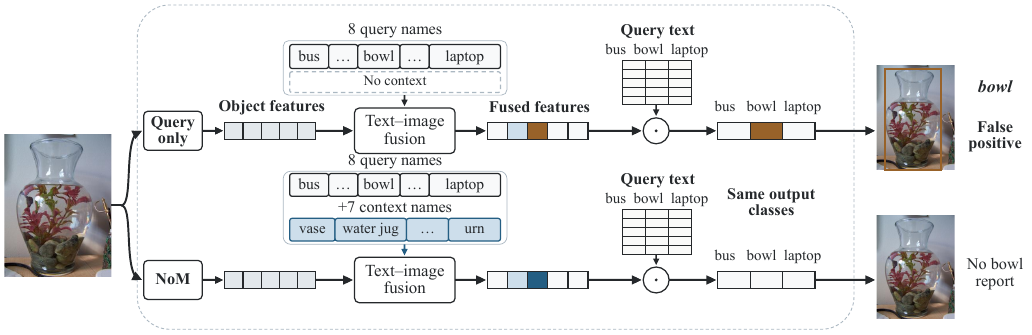}
\caption{\textbf{Omitting a scene name can turn an unnamed object into a queried false positive.} Evaluation usually treats the class list only as the set of allowed answers. We find that in text-fusion detectors it also conditions the evidence the detector computes. With eight queried classes (\mbox{GD-L}, COCO 298904), the query-only prompt reports the unnamed vase as \emph{bowl} (0.294). NoM adds seven context names and removes this report without adding any output label. Other reports are omitted, and feature and score strips are schematic.}
\label{fig:phenomenon}
\end{figure*}

A user looking for bowls should be able to ask for bowls without naming the vase beside them. Open-vocabulary detectors accept such queries~\cite{glip,groundingdino,mmgd,llmdet,yoloworld,minderer}, yet benchmarks typically supply the dataset vocabulary, either in full or in blocks~\cite{lvis,wedetect}. In detectors that fuse text into visual features, a name helps shape the representation of the scene even when the user has no interest in reporting that class. Remove \emph{vase} and the detector may report the vase as another bowl (Fig.~\ref{fig:phenomenon}). On COCO, Grounding DINO-L (\mbox{GD-L}) loses 16.2 AP50 points when evaluated on the same eight classes with a short prompt. Even the ordering of detectors can reverse. We call this loss the \emph{omission cost}.

One explanation concerns how the detector distributes attention. Tractor-Dogs~\cite{tractordogs} showed that a detector can report a queried class even when it is absent and introduced extra prompt tokens, called \emph{attention sinks}, that can absorb attention. It also showed that naming a present class can help. In the bowl example, a sink and the word \emph{vase} can receive equal shares of a region's attention. Should they have the same effect on its bowl score?

Our experiments show that they need not. To isolate the role of the name, we keep the labels the detector can report fixed and add equal numbers of present or absent class names to its input. Present names recover much more accuracy. With the classifier fixed, withdrawing an object's own name makes a competing label more likely at its location. We call this error a \emph{takeover}. An identity for a single attention head explains why the choice of name matters. The attention weight scales the change in the head's output. The direction of that change relative to the competing score's gradient determines whether the score rises or falls to first order. A name closely related to a queried class can even depress the score of a genuine instance of that class.

These findings motivate NoM. A frozen image model proposes scene names. NoM removes names closely related to the query and supplies the rest as context. At final scoring, or \emph{readout}, only the user's classes can be reported. We call this separation \emph{context--answer decoupling}. NoM requires no training. Under the same output restriction, NoM outperforms attention sinks on \mbox{GD-L} and \mbox{LLMDet-L} and recovers most of the omission cost (Sec.~\ref{sec:main}). Its gain tracks the omission cost across checkpoints. We also test the two roles of the list in a generative detector (Sec.~\ref{sec:generative}).

Our contributions are as follows.
\begin{itemize}
\item We measure the omission cost on the same images and queried classes and show how full-vocabulary evaluation can reverse the ranking observed under partial queries (Sec.~\ref{sec:measurement}).
\item We trace much of this cost to missing names of present objects, locate where their removal changes recognition, and derive a signed first-order prediction of each object's score change (Sec.~\ref{sec:measurement}).
\item We introduce NoM and evaluate its name proposals and filtering under a common output restriction. We show where it improves on attention sinks, where added context interferes, and how the roles of the class list can diverge in a generative detector (Secs.~\ref{sec:nom}--\ref{sec:limits}).
\end{itemize}

\section{Related Work}
\paragraph{Text conditioning and visual naming.}
GLIP, Grounding DINO, MM-Grounding-DINO, LLMDet, and \mbox{YOLO-World} use text conditioning~\cite{glip,groundingdino,mmgd,llmdet,yoloworld}. OWLv2 and WeDetect supply text-independent visual references~\cite{minderer,wedetect}. We intervene on implemented text paths without inferring invariance from architecture labels. Munim et al.~\cite{munim2026calibrate} show that candidate-dependent scores break conformal risk calibration as the vocabulary grows and restore it with a fixed conditioning bank. We hold the evaluated classes fixed as other names are omitted and measure detection accuracy. Generative detectors such as \mbox{Qwen2.5-VL}~\cite{bai2025qwen} produce boxes as text, so the class list can reach both the image states and the readout.

Understand and Detect generates image-specific prompts using foreground extraction, image--text pairs, and adaptive scoring~\cite{understanddetect}. It establishes precedent for automatic visual naming. In NoM, $Q$ denotes the user's fixed set of queried classes. Image-proposed names supply additional context. The proposed names only condition the detector, never become eligible labels, and are screened against $Q$ because their local score response can turn negative. CoCoOp and PDP learn conditional prompts or prompt pools~\cite{cocoop,pdp}, and Grounded SAM and SAM~3 connect naming with grounding and segmentation~\cite{groundedsam,sam3}.

\paragraph{Fusion errors and attention sinks.}
Tractor-Dogs studies false reports when no prompted object is present, traces them to vision--language fusion, and shows that present-class context can help~\cite{tractordogs}. We measure recognition of the same queried classes as other scene names are omitted. NoM supplies image-proposed context while keeping the answer set fixed. In the original sink method, a sink can receive the highest class score, in which case the prediction is discarded. In our comparison, the detector chooses among the queried classes for both methods. We compare query-class AP and takeovers of present, unnamed objects.

\paragraph{Unknown objects and controlled analysis.}
Open-world detectors study discovery and reporting beyond known classes~\cite{joseph,gupta,prob,owobj,uniow,owel,noovd}. Our main endpoint is query-class recognition. Removing a name during fusion while holding the classifier fixed is related to model mediation and attribution patching~\cite{vig,atpstar}. Same Attention, Different Truths studies attention and hallucinated mentions in generated language~\cite{sameattention}. Our signed response instead predicts a specified detector-score intervention, and NoM is evaluated separately on detection AP.

\section{Missing Names Change Recognition}
\label{sec:measurement}
We first ask how much a partial query costs, then what carries the cost and why some names help while others hurt.

\begin{figure*}[t]
\centering\includegraphics[width=.86\textwidth]{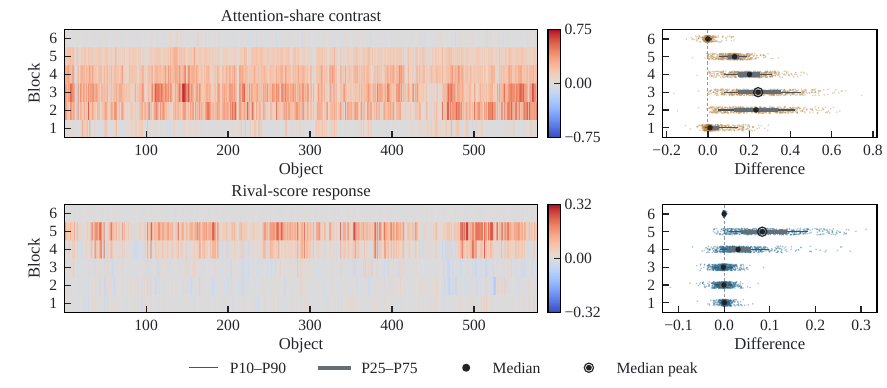}
\caption{\textbf{Attention and response peak at different blocks.} GLIP-T(B), 577 objects in the same column order across six blocks. Top: baseline own-minus-absent attention share. Bottom: final score contrast from own- versus absent-name withdrawal under fixed readout. Right: all objects, medians, and percentile ranges (not confidence intervals). Median peaks: blocks 3/5. Color scales differ.}
\label{fig:layer_objects}
\end{figure*}

\begin{figure*}[t]
\centering\includegraphics[width=\textwidth,trim=0bp 1bp 0bp 1bp]{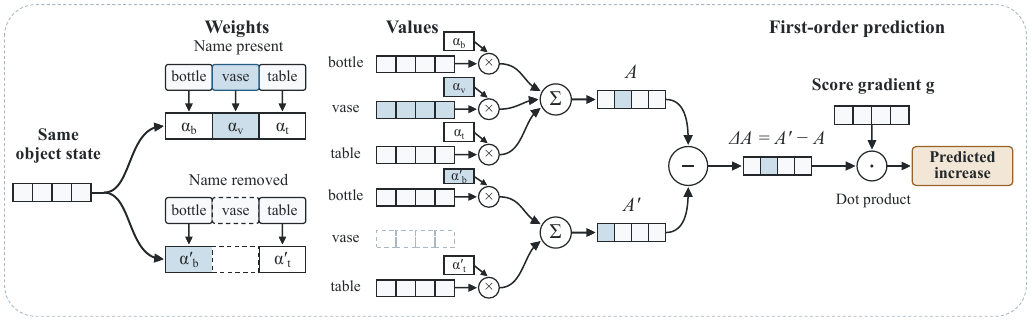}
\caption{\textbf{Withdrawing a name changes both the content and the normalization of attention.} Each value contributes to $A$ or $A^{\prime}$ with the labeled weight. The net change $\Delta A=A^{\prime}-A$ is projected onto the competing-score gradient $g$. The illustrated response raises the competing score, and its sign can reverse. Original local states are fixed, and the score prediction is first order.}
\label{fig:local_operation}
\end{figure*}

\subsection{Partial queries cost accuracy and reorder detectors}
For a fixed set of images and queried classes $Q$, we measure omission cost as the difference between full-prompt and short-prompt AP50 (average precision at an intersection-over-union threshold of $0.5$). The full-prompt reference scores all COCO classes, applies the detector's score threshold and limit on the number of reports, then retains predictions in $Q$. This comparison captures the cost a user experiences when shortening the prompt. The interventions below isolate the effect of changing the text used inside the detector.

LLMDet leads WeDetect with the full vocabulary, but WeDetect takes the lead with eight queried classes. The loss varies even among detectors that use text during visual processing. HDINO-T~\cite{hdino2026} and OV-DEIM-S~\cite{ovdeim2026} show no positive omission cost, and the detectors with text-independent visual features, including YOLOE~\cite{yoloe2025}, change little. Supplementary Table~\ref{tab:rankscale} gives the rankings and paired costs by checkpoint. Results by query size and detector family appear in Suppl.~Sec.~\ref{app:a1_details}.

\begin{figure*}[t]
\centering\includegraphics[width=.86\textwidth]{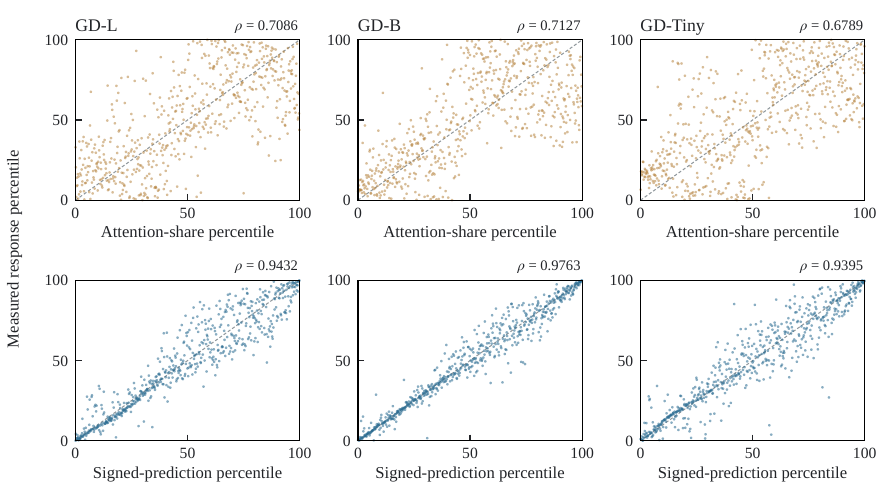}
\caption{\textbf{Direction predicts withdrawal responses better than attention share.} All 587 objects per checkpoint. Axes show predictor and response percentile ranks. Top: own-name attention share. Bottom: signed first-order prediction. $\rho$: Spearman correlation. The diagonal denotes equal ranks, not equal raw score changes.}
\label{fig:direction_full}
\end{figure*}

\subsection{Names of present objects carry most of the lost context}
What information does the short prompt lose? We use annotations to add names of objects present in the image or the same number of names of absent classes. In both conditions the detector can report only the original query classes. On \mbox{GD-L}, present names recover 12.43 AP50 points over the short prompt. Absent names yield little improvement. LLMDet-L shows the same pattern, and on both detectors the full vocabulary still supplies some additional benefit (Suppl.~Fig.~\ref{fig:present} and Table~\ref{tab:oracle}).

With the full prompt held fixed, restricting the classifier to $Q$ gives nearly the same accuracy as letting it choose among all classes and then discarding reports outside $Q$. These comparisons locate most of the omission cost on these detectors in the information available while processing the image. They motivate proposing scene names from the image when annotations are unavailable.

\subsection{Removing an object's own name changes its representation}
To locate the effect, we remove the text tokens for an annotated object's name at a chosen fusion layer. As a control, we remove the tokens of a matched absent class at the same layer. Both interventions keep the classifier text and competing labels fixed and exclude the object's own class from the possible answers. We measure whether a competing prediction covers the target at intersection over union greater than $0.5$. The difference in takeover probability is

\begin{equation}
\Delta_{\rm fixed}=\E[Y_{\rm own,fixed}-Y_{\rm absent,fixed}],
\label{eq:fixed}
\end{equation}
where $Y$ records whether a competing prediction covers the target. Box geometry is fixed for GLIP, YOLO-World, and OWLv2 and can respond to the intervention in the Grounding DINO and HDINO experiments.

In \mbox{GD-L}, withdrawing the object's name has its largest measured effect at \emph{decoder text attention}, where candidate regions attend to the prompt before classification. Takeover probability rises by 15.84 percentage points relative to withdrawing an absent name. Intervening earlier, during feature enhancement or candidate selection, has a much smaller effect. GLIP is most sensitive in its late fusion blocks, even though attention to the object's name peaks earlier (Fig.~\ref{fig:layer_objects}). YOLO-World also responds with its boxes fixed. In the text-independent OWLv2 control, the intervention leaves the visual features and scores unchanged. Supplementary Sec.~\ref{app:score_accounting} gives the protocol and comparisons by site.

\begin{figure*}[t]
\centering\includegraphics[width=\textwidth,trim=0bp 1bp 0bp 1bp]{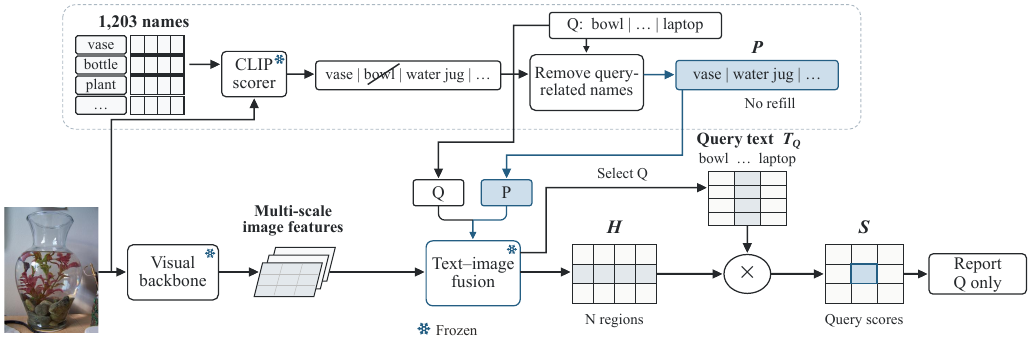}
\caption{\textbf{NoM conditions on scene names but reports only the query.} CLIP selects eight of 1,203 names. A fixed concept rule removes query relatives without refilling, and a fitting prefix enforces the 256-token budget. The image and $Q\cup P$ enter the frozen detector, and region representations $H$ are scored only against the query columns $T_Q$. Matrices and candidate names are illustrative.}
\label{fig:nom_operation}
\end{figure*}

\subsection{The direction of a name's update predicts its score effect}\label{sec:local}
Present names help on average, yet an added name can also lower a queried score. To see why, we analyze one attention operation. A name contributes through both its value and its share of the attention normalization (Fig.~\ref{fig:local_operation}). Consider one attention head, with query $q$, keys $k_j$, and output-projected values $u_j=W_ov_j$. Write $p_j=\exp(\ell_j)/Z_Q$ for the original distribution over prompt tokens, $\ell_j=q^\top k_j/\sqrt d$, and $A_Q=\sum_{j\in Q}p_ju_j$. Insert a set $P$ while holding the original query, key, and value states fixed. With $Z_P=\sum_{j\in P}\exp(\ell_j)$, $m_P=Z_P/(Z_Q+Z_P)$, and $\bar A_P=\sum_{j\in P}\exp(\ell_j)u_j/Z_P$, we obtain
\begin{align}
A_{Q\cup P}&=(1-m_P)A_Q+m_P\bar A_P,\label{eq:insert}\\
\delta A&=-m_PA_Q+m_P\bar A_P.\label{eq:two}
\end{align}
For withdrawal of a span $S$ carrying original mass $a_S<1$, the corresponding identity is
\begin{equation}
A_{\setminus S}-A=\frac{a_S}{1-a_S}(A-\bar A_S).
\label{eq:remove}
\end{equation}
These identities are exact for one attention operation with fixed original states. Changes that a new text input causes in earlier layers are additional paths outside Eq.~\eqref{eq:insert}.

Let $g=\partial s_c/\partial A$ for a fixed candidate and class. By Eq.~\eqref{eq:two}, the first-order score prediction is
\begin{equation}
g^\top\delta A=m_P\bigl(g^\top\bar A_P-g^\top A_Q\bigr).
\label{eq:rule}
\end{equation}
This \emph{direction rule} separates mass from orientation: $m_P>0$ scales the update, but $s_c$ rises to first order only when inserted values project onto $g$ more strongly than the current output. Attention alone cannot determine the sign. \mbox{Equation~\eqref{eq:remove}} gives the withdrawal version. Approximation error and subsequent geometric changes in nonlinear detectors must be measured separately.

We compare the two predictions on the same objects after withdrawing each object's full name. Across the Grounding DINO checkpoints, the signed prediction consistently tracks the measured change in the competing score more closely than the attention share on the name (Fig.~\ref{fig:direction_full}). On \mbox{GD-L} its rank correlation with the measured change is $0.9432$. Supplementary Sec.~\ref{app:added_evidence} gives the paired differences and intervals.

\paragraph{Implication for context selection.}
A proposed name can describe something in the image and still lower the score of a queried object. For example, related train names can depress the score of a genuine train. On the separate selection set, Eq.~\eqref{eq:rule} predicts a larger drop for true queried objects from names excluded by NoM's concept filter than from names it retains (Suppl.~Table~\ref{tab:fu_insert}). This motivates a fixed semantic filter whose effect on accuracy we test by ablation. NoM requires no gradients at inference.

\section{NoM Separates Context from Answers}
\label{sec:nom}
Section~\ref{sec:measurement} found that names of present objects carry most of the loss, that a name acts by changing the representations the detector computes, and that the direction of its update, not its attention mass, sets whether it helps. NoM turns these findings into three steps (Fig.~\ref{fig:nom_operation}). It proposes scene names from the image and removes names related to the query (Sec.~\ref{sec:nom_names}), and it restricts the readout to the user's classes so that the added names condition the detector without becoming answers (Sec.~\ref{sec:nom_readout}). The detector stays frozen, and nothing is trained.

\subsection{Nominate scene names and screen query relatives}\label{sec:nom_names}
Section~\ref{sec:measurement} shows that present names carry most of the loss, so NoM must supply them without annotations. A frozen CLIP image encoder~\cite{clip} ranks 1,203 LVIS names by image--text similarity. From the top eight, we remove exact query duplicates and query-related concepts without refill, then retain the longest prefix fitting the 256-token budget. The proposal count and single-pass recipe were selected on a separate 2,000-image set using both detectors before validation scoring. End-to-end latency includes the extra image-model pass.

Section~\ref{sec:local} shows that presence need not imply a helpful update under $Q$. We screen query relatives with a fixed concept relation $\mathcal R(q,n)$: WordNet 3.0~\cite{wordnet} synonymy, directed hypernym/hyponym relations, direct part/member/substance relations, or complete-word-sequence containment. Applying it to the original proposal list gives
\begin{equation}
P=\{n\in P_{\rm raw}:\ \neg\mathcal R(q,n)\ \forall q\in Q\}.
\label{eq:m1}
\end{equation}
NoM denotes the filtered recipe, and raw names is its unfiltered baseline. The paired comparison fixes the image, CLIP scorer, original top-eight list, and detector to isolate filtering. We also test the separately fixed 24-sink list with NoM.

\subsection{Condition on more names, report only the query}\label{sec:nom_readout}
The added names must condition the detector without entering the answer set, as in the controls of Sec.~\ref{sec:measurement}. The frozen detector processes image $x$ with $Q\cup P$ (Fig.~\ref{fig:nom_operation}), producing states $H(x,Q\cup P)$ and boxes $B(x,Q\cup P)$. Its readout uses only $Q$:
\begin{equation}
s_{ic}=f_c(H_i(x,Q\cup P),T_c),\qquad c\in Q.
\label{eq:interface}
\end{equation}
Names in $P$ can change fused features but cannot become reported labels. This is context--answer decoupling. The same output restriction is used for the raw-name and sink controls. 

Under \emph{Q-only readout}, each candidate chooses among $Q$. In \emph{open readout then filtering}, it chooses among a larger vocabulary before non-query reports are discarded. The latter can suppress queried reports through label competition alone. Primary comparisons use Q-only readout. Historical implicit-object readouts are kept separate.
NoM requires one conditioned detector pass after nomination. Scores, boxes, and ranking can change despite the fixed output set. We next test recovery and the filter's incremental value.

\section{NoM Recovers Query-Class Accuracy}
\label{sec:main}
Sections~\ref{sec:measurement} and~\ref{sec:nom} predict that image-proposed names, screened by the concept rule, should recover much of the omission cost while the answer set stays fixed. This section tests that prediction.

\begin{table*}[t]
\centering\small\setlength{\tabcolsep}{2.5pt}
\caption{\textbf{NoM recovers most of the omission cost on the unchanged query task.} COCO val2017, 5,000 images. AP parentheses show gains over native. Bold method names mark our variants. Bold/underlined numbers mark best/second among the displayed corrections within a detector. References and timing are unranked. NoM w/o filter is Raw names. The tuned fixed-word reference is in Suppl.~Table~\ref{tab:additional_paired_N}. Corrections use query-only output. TO8/100: takeovers per 100 unqueried GT (lower is better). $E$: end-to-end with nomination, $\Sigma$: stage sum. Full80$\to$Q8 selects among all names, then filters. WeDetect uses native resources (Suppl.~Table~\ref{tab:rankscale}). Dashes mark unmeasured cells.}
\label{tab:main}
\begin{tabular}{llcccccc}\toprule
Model & Method & Q8 AP ($\Delta$) & Q8 AP50 & Q16 AP50 & full80 AP & TO8/100 & $t_{\rm Q8}$ (ms)\\\midrule
\mbox{GD-L}~\cite{groundingdino} & Native short &44.26\;(+0.00)&55.78&57.61&\textbf{52.60}&52.09&557.1\textsuperscript{$E$}\\
 & 24 sinks (adapted)~\cite{tractordogs} &49.08\;(+4.82)&63.77&63.46&\underline{51.48}&24.57&584.3\textsuperscript{$E$}\\
 & CLIP veto~\cite{clip} &40.47\;($-$3.79)&51.78&53.79&47.93&41.15&727.09\textsuperscript{$\Sigma$}\\
\cmidrule(lr){2-8}
 & \textbf{NoM w/o filter (ours)} &\underline{51.67}\;(+7.41)&\underline{68.23}&\underline{66.19}&50.22&\underline{10.44}&616.82\textsuperscript{$\Sigma$}\\
 & \textbf{NoM (ours)} &\textbf{52.18}\;(+7.92)&\textbf{68.58}&\textbf{66.94}&50.73&10.59&600.5\textsuperscript{$E$}\\
 & \textbf{NoM + 24 sinks (ours)}~\cite{tractordogs} &51.59\;(+7.33)&67.78&--&50.54&\textbf{6.92}&636.8\textsuperscript{$E$}\\
\cmidrule(lr){2-8}
 & Full80 $\to$ Q8 ref. &55.60&71.99&--&--&--&--\\
\midrule
\mbox{LLMDet-L}~\cite{llmdet} & Native short &51.21\;(+0.00)&63.57&66.06&\underline{59.00}&58.30&541.2\textsuperscript{$E$}\\
 & 24 sinks (adapted)~\cite{tractordogs} &53.63\;(+2.42)&66.67&67.86&\textbf{59.04}&45.22&569.9\textsuperscript{$E$}\\
 & CLIP veto~\cite{clip} &48.30\;($-$2.91)&60.59&62.38&54.16&47.19&724.22\textsuperscript{$\Sigma$}\\
\cmidrule(lr){2-8}
 & \textbf{NoM w/o filter (ours)} &55.20\;(+3.99)&70.07&\underline{69.39}&57.21&\underline{24.35}&592.67\textsuperscript{$\Sigma$}\\
 & \textbf{NoM (ours)} &\underline{56.94}\;(+5.73)&\underline{72.15}&\textbf{70.63}&58.00&24.83&579.5\textsuperscript{$E$}\\
 & \textbf{NoM + 24 sinks (ours)}~\cite{tractordogs} &\textbf{57.61}\;(+6.40)&\textbf{72.97}&--&58.45&\textbf{20.53}&612.9\textsuperscript{$E$}\\
\cmidrule(lr){2-8}
 & Full80 $\to$ Q8 ref. &60.79&76.50&--&--&--&--\\
\midrule
\multicolumn{2}{l}{WeDetect~\cite{wedetect}, native Q8 ref.} &55.46&73.05&--&--&--&--\\
\bottomrule
\end{tabular}
\end{table*}

\subsection{Evaluation setup}
We evaluate all 5,000 COCO val2017 images~\cite{coco}. Q8, Q16, and full80 query 8, 16, and all 80 classes. Q8 is primary, and full80 tests complete prompts. Grounding DINO-family comparisons use 900 candidates, one label each, at most 300 reports per image, and score $>0.05$. Image-hash queries are annotation-independent. The NoM configuration was fixed on a separate 2,000-image selection set before COCO validation scoring, and all its rules, thresholds, and selection protocols are specified in Sec.~\ref{sec:nom} and the supplementary material, so NoM can be reimplemented without training.

AP averages IoUs $0.50{:}0.05{:}0.95$, and AP50 uses $0.50$. Unless noted, intervals use 5,000 paired image bootstrap resamples, grouping all objects from each image. Checkpoints sharing images are not independent physical samples. Timing includes nomination. We specified the query draws, alternative image-based proposer, and evaluations on LVIS, \mbox{ODinW-13}, and shorter queries before observing their outcomes (Suppl.~Secs.~\ref{app:robustness}--\ref{app:further}). Configurations and reporting rules: Suppl.~Secs.~\ref{app:a1_details}, \ref{app:expanded_details}. Detailed ablations: Suppl.~Secs.~\ref{app:additional_controls}--\ref{app:n2}.

Controls are random names, fixed names plus sinks, and CLIP veto. The 24 mean-initialized sinks traverse the full detector but are excluded before argmax: this is our query-task adaptation, not the original Tractor-Dogs benchmark. Error definitions and reporting details appear in Suppl.~Secs.~\ref{app:visual_comparisons} and~\ref{app:expanded_details}.

\begin{table}[tbp]
\centering\small\setlength{\tabcolsep}{3pt}
\caption{\textbf{Core component ablation in the main query task.} Q8 AP on the same 5,000 images. N: image-proposed names, F: concept filtering, S: 24 sinks. The detector and reporting settings are fixed. Bold method names identify our variants. Bold/underlined numbers mark best/second in each column.}
\label{tab:core_ablation}
\begin{tabular}{lcccrr}\toprule
Variant & N & F & S & \mbox{GD-L} & \mbox{LLMDet-L}\\\midrule
Native & -- & -- & -- & 44.26 & 51.21\\
Sinks~\cite{tractordogs} & -- & -- & $\checkmark$ & 49.08 & 53.63\\
\textbf{w/o filter} & $\checkmark$ & -- & -- & \underline{51.67} & 55.20\\
\textbf{NoM} & $\checkmark$ & $\checkmark$ & -- & \textbf{52.18} & \underline{56.94}\\
\textbf{NoM + sinks} & $\checkmark$ & $\checkmark$ & $\checkmark$ & 51.59 & \textbf{57.61}\\\bottomrule
\end{tabular}
\end{table}

\begin{figure*}[t]
\centering\includegraphics[width=.90\textwidth]{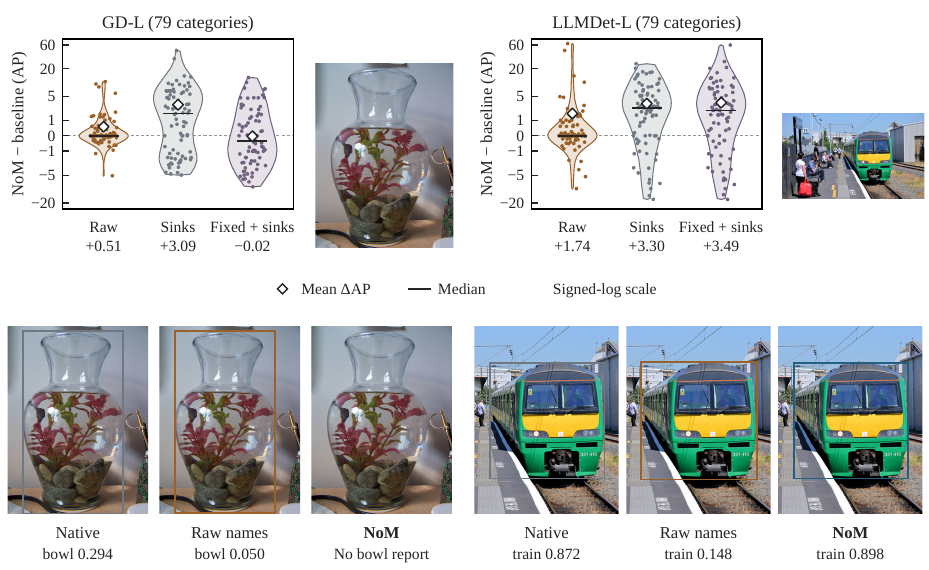}
\caption{\textbf{NoM gains vary across classes and baselines.} Top: Q8 AP differences (NoM minus baseline), all 79 estimable classes. Diamonds/lines mark means/medians. Signed-log axes retain AP units. Bottom: stored native/raw/NoM predictions. Unrelated reports omitted. Context removes a false bowl report (left). Filtering restores a train report (right).}
\label{fig:class_differences}
\end{figure*}

\subsection{NoM recovers most of the omission cost}
NoM recovers most of the omission cost in both detectors (Table~\ref{tab:main}, Fig.~\ref{fig:class_differences}). At Q8, its standard-AP gains over native are $7.92$ and $5.73$ points, and its gains over the same-readout sink baseline are $3.10$ and $3.31$. AP50 also improves: by $12.80/8.58$ over native and $4.82/5.48$ over sinks. NoM also comes within 0.02 AP of a development-tuned fixed-word-plus-sink reference on \mbox{GD-L} ($-0.02$ $[-1.03,0.74]$) and exceeds it on \mbox{LLMDet-L} by $3.49$ AP and $5.38$ AP50. Only image-proposed context helps both detectors without per-detector tuning (Suppl.~Table~\ref{tab:additional_paired_N}).

Raw names also exceed equal-count random names and CLIP veto at Q8. GD-Tiny and GLIP transfer contrasts, including negative endpoints, are in Suppl.~Sec.~\ref{app:control_details}.

WeDetect~\cite{wedetect}, the strongest native checkpoint at Q8, reaches 55.46 AP and 73.05 AP50. \mbox{LLMDet-L} with NoM reaches 56.94 AP, a paired gain of $1.48$ $[0.04,3.17]$ (Suppl.~Sec.~\ref{app:added_evidence}). Its AP50 difference is $-0.90$ $[-2.61,0.97]$, an interval that includes zero. Before filtering, the raw-name AP50 gap was $2.98$ $[0.99,4.76]$. This reference uses different resources: WeDetect has 33,600 candidates and the authors' Chinese text bank, versus 900 candidates in the GD family (Suppl.~Sec.~\ref{app:mn}).

\paragraph{Variation across query draws.}
NoM fits no parameters. Across three fixed query draws (original plus two hash salts), Q8 gains over native are $12.99\pm0.17$ AP50 for \mbox{GD-L} and $8.37\pm0.23$ for \mbox{LLMDet-L} (mean $\pm$ sample SD). All draws: Suppl.~Fig.~\ref{fig:draw_plot}.

\subsection{Component ablation}
Table~\ref{tab:core_ablation} isolates the core components under the same Q8 task. Concept filtering improves standard AP on both detectors at Q8, Q16, and full80 (Suppl.~Table~\ref{tab:mn}). At Q8, gains over the same-input raw names are $0.51$ $[0.16,0.96]$ AP for \mbox{GD-L} and $1.74$ $[0.97,2.51]$ for \mbox{LLMDet-L}. With the proposal list fixed, AP50 gains are $0.36$ $[-0.06,0.83]$ and $2.09$ $[1.21,2.82]$ (Suppl.~Table~\ref{tab:mn}). At Q16, AP50 increases by $0.75$ $[0.26,1.26]$ and $1.25$ $[0.79,1.74]$, and at full80 by $0.49$ $[0.28,0.74]$ and $0.82$ $[0.66,0.97]$. Filtering also reduces the complete-prompt cost.

The two operations serve different roles. Raw names reduce Q8 takeover from $52.09$ to $10.44$ per 100 unqueried GT in \mbox{GD-L} and from $58.30$ to $24.35$ in \mbox{LLMDet-L}. Filtering then restores queried reports (Fig.~\ref{fig:class_differences}): its Q8 recovered/lost counts are $7/4$ and $9/1$, while takeover rates change little (Suppl.~Table~\ref{tab:mn}).

\section{When Context Helps and When It Interferes}
\label{sec:limits}
If NoM works by restoring missing context, its gain should shrink when little is missing, and added context can interfere when nothing is missing. This section tests these predictions, compares names with sinks, and examines the same two roles of the class list in a generative detector.
\subsection{Benefits shrink with query coverage}

With the raw-name recipe held fixed, its gain diminishes as more queried names are supplied (Suppl.~Fig.~\ref{fig:curve}). Full-prompt conditioning followed by filtering remains stronger. With all names supplied, raw names hurt both models. The exact paired contrasts and standard-AP changes are retained in Suppl.~Sec.~\ref{app:a4_details}.

\subsection{Names and sinks are complementary}

Adding names and sinks together reduces the recorded errors further than either alone in \mbox{GD-L} and \mbox{LLMDet-L}. Supplementary \mbox{Figure~\ref{fig:error_matrix}} uses the same concept-filtered NoM throughout, and all 18 adjacent paired 95\% intervals decrease. Both additions reduce both errors, and the intervals are per contrast.

Combining names and sinks provides the lowest recorded error rates, while AP depends on the full precision--recall ranking. At Q8, adding sinks to NoM changes AP50 by $-0.81$ $[-1.45,-0.17]$ in \mbox{GD-L} and $+0.82$ $[0.23,1.39]$ in \mbox{LLMDet-L}. The combination suppresses errors in both models and improves AP50 in \mbox{LLMDet-L}, so both recipes remain in the main table. At full80, reserving sink tokens truncates proposed names in 4,320/5,000 images, making that comparison a joint context change.

\paragraph{Removing benchmark names from the proposal bank.}
Deleting the 78 directly represented COCO class names from the proposal bank lowers NoM Q8 AP50 by $0.99$ $[0.41,1.65]$ in \mbox{GD-L} and $0.89$ $[0.55,1.42]$ in \mbox{LLMDet-L}. The remaining AP50 values are $67.59$ and $71.26$. Most of the gain survives removal of those exact or direct-synonym entries.

\subsection{Transfer to a restricted open-world vocabulary}
On the M-OWODB test set, which was also used during development, we use frozen models, a legal 874-name bank excluding registered future classes, and the authors' 100-report budget with score $>0.05$. All variants report known classes only, so unknown AP and recall are zero. NoM reduces A-OSE at both 20- and 40-known-class settings, but recognition responds differently: known AP50 changes by $+0.97/+1.30$ in \mbox{GD-L} and $-3.63/-1.87$ in \mbox{LLMDet-L}. The object-token reference has higher \mbox{GD-L} AP50 at both stages. Exact values, paired intervals, object accounting, and all legal-bank variants are retained in Suppl.~Sec.~\ref{app:ow_details} and Suppl.~Fig.~\ref{fig:owod_plot}.

\subsection{When NoM is not needed}
Three settings leave little for NoM to recover.
\paragraph{A complete prompt.}
The same operation remains a perturbation when the query is complete. With all 80 names supplied, NoM loses $2.33$ and $0.91$ AP50 in \mbox{GD-L} and \mbox{LLMDet-L}. Extra conditioning can interfere with a complete prompt, even with a frozen detector. A development-only decoder gate also leaves a residual cost (Suppl.~Sec.~\ref{app:oracle_gate}).
\paragraph{Context quality and checkpoint dependence.}
Across the eight text-conditioned checkpoints, the Q8 AP50 gain of NoM tracks the omission cost (Spearman 0.83, Suppl.~Fig.~\ref{fig:cost_gain}). NoM helps where the cost is large and is unnecessary where little is missing, as on the Grounding DINO-B checkpoint fine-tuned on COCO ($-4.81$ $[-6.14,-3.47]$ AP50) and \mbox{OV-DEIM} ($-1.62$ $[-2.27,-0.69]$). GD-Tiny shows that proposal quality also matters: raw names give no confirmed Q8 AP50 gain there despite about 79\% oracle recovery. Matched learned pools and confirmation scorers do not improve on the fixed recipe (Suppl.~Sec.~\ref{app:additional_controls}).
\paragraph{Text-independent models.}
Their visual states have no missing-name shift for NoM to repair, although a changed label set can still change output winners.

\subsection{The class list in a generative detector}
\label{sec:generative}
A generative detector shows the same two roles of the class list in a different balance. We use \mbox{Qwen2.5-VL-7B}~\cite{bai2025qwen} with greedy decoding constrained to a JSON list of boxes whose labels must come from the eight queried classes, and score micro-F1 at IoU 0.5 (Suppl.~Sec.~\ref{app:generative}). Because attention is causal, image states can depend on the list only when the list precedes the image. With the image first, Q8 and full-list prompts give identical image states in all 29 compared layers. With the list first, the states differ from the first decoder layer on.

With the list first, we replay in every decoder layer the image states of each image's Q8 run into its full-list run, leaving the visible list and the allowed labels unchanged (Table~\ref{tab:qwen}). On 1,900 COCO images not used in earlier runs, the full list lowers F1 from 37.46 to 28.31, by 9.16 $[5.68,12.83]$ points. With Q8 image states it lowers F1 by 11.67 $[8.39,15.27]$ points, so the loss arises at the readout. Full-list image states recover 2.51 $[0.25,4.81]$ of these points, the registered primary contrast. The signs match those of text-fusion detectors: conditioning on more names helps, and names present when the answer is produced hurt. The balance differs. In \mbox{GD-L} and \mbox{LLMDet-L}, Q-only full conditioning adds 16.56 and 13.14 AP50 (Suppl.~Table~\ref{tab:oracle}), while open readout costs only 0.35 and 0.21 (Sec.~\ref{sec:measurement}). In \mbox{Qwen2.5-VL}, the readout effect is more than four times the conditioning effect. This also accounts for an earlier registered prediction that was not met: on 500 images with free decoding, the full list did not raise F1 when it preceded the image (Suppl.~Sec.~\ref{app:generative}).

\begin{table}[t]
\centering\small\setlength{\tabcolsep}{4pt}
\caption{\textbf{The class list in \mbox{Qwen2.5-VL}.} List before image, 1,900 COCO images. Every arm allows only the eight queried labels. Replayed image states come from the same image's Q8 run in every decoder layer. Micro-F1 at IoU 0.5, paired 95\% intervals.}
\label{tab:qwen}
\begin{tabular}{llrrr}\toprule
Image states & List at readout & F1 & Matched & Reports\\\midrule
Q8 & Q8 & \textbf{37.46} & 564 & 1,430\\
Full 80 & Full 80 & 28.31 & 429 & 1,450\\
Q8, replayed & Full 80 & 25.79 & 398 & 1,505\\\midrule
\multicolumn{2}{l}{Full list, total change} & \multicolumn{3}{r}{$-9.16$ $[-12.83,-5.68]$}\\
\multicolumn{2}{l}{Readout only} & \multicolumn{3}{r}{$-11.67$ $[-15.27,-8.39]$}\\
\multicolumn{2}{l}{Full-list image states} & \multicolumn{3}{r}{$+2.51$ $[0.25,4.81]$}\\\bottomrule
\end{tabular}
\end{table}

\section{Conclusion}
The class list of an open-vocabulary detector is usually treated as the set of allowed answers. For detectors that fuse text into visual features, it also conditions the evidence the detector computes. Partial queries change internal representations, cost up to 16.2 AP50 with eight queried classes and 19.4 with one, and reorder detectors. The missing names of present objects carry most of this cost, and the direction of a name's update, not the attention it receives, decides whether the name helps or hurts. NoM recovers most of this omission cost without training by letting the detector see more names than it may report and screening those whose change can point the wrong way, and the measured cost tells when to apply it. In \mbox{Qwen2.5-VL} the same two roles of the class list separate, and the readout effect dominates. Open-vocabulary benchmarks should report partial-query accuracy alongside full-vocabulary accuracy and record the names used for conditioning, the names visible at the readout, and the allowed answers.

\clearpage
\section*{LLM Usage Statement}
OpenAI Codex assisted with refining author-produced experiment and evaluation code, checking literature and references, reviewing data analyses and mathematical derivations, and editing the manuscript. The authors take full responsibility for all submitted content, regardless of any LLM or AI-tool use.

{\small\bibliographystyle{ieeenat_fullname}\bibliography{main}}

\clearpage
\renewcommand{\thepage}{S\arabic{page}}
\setcounter{page}{1}
\maketitlesupplementary
\appendix

\renewcommand{\floatpagefraction}{0.90}
\renewcommand{\dblfloatpagefraction}{0.90}

\hyphenation{Table Tables Figure Figures Section Sections Equation}
\newcommand{\supports}[1]{\noindent\textit{Supports Sec.~\ref{#1}.}\ }
\section*{Overview}
\noindent\textbf{NoM at a glance.} In a text-fusion detector the class list is both the answer set and a conditioning input. NoM separates the two roles, which the main paper calls context--answer decoupling (Sec.~\ref{sec:nom}). The frozen detector is conditioned on the query $Q$ plus image-proposed names $P$ but reports only classes in $Q$. Algorithm~\ref{alg:nom} lists the three steps, and Table~\ref{tab:steps} links each step to the measurement that requires it and to the evidence that it helps. NoM trains no parameter, uses no annotation at inference, and adds about 40\,ms per image including nomination (Table~\ref{tab:main}). The gain holds across five query draws, four proposers, and three nomination budgets on both detectors, reaches 13.91 and 10.54 AP50 with a single queried class, and, for \mbox{GD-L}, transfers to a fixed eight-class LVIS query and to four of seven multi-class ODinW domains (Secs.~\ref{app:robustness} and~\ref{app:odinw}).

\newcounter{savedtable}\setcounter{savedtable}{\value{table}}\setcounter{table}{0}
\let\origthetable\thetable\let\origtheHtable\theHtable
\renewcommand{\thetable}{\Roman{table}}\renewcommand{\theHtable}{overview.\arabic{table}}
\begin{algorithm}[!h]
\caption{NoM inference. No parameter is trained.}
\label{alg:nom}
\begin{algorithmic}[1]
\Require image $x$, query classes $Q$, frozen text-fusion detector $D$, frozen CLIP encoder, name bank $V$ (1,203 LVIS names), $k=8$, token budget 256
\Statex \textit{Step 1: scene nomination (Sec.~\ref{sec:nom})}
\State $P_{\rm raw}\gets$ the $k$ names in $V$ most similar to $x$ under CLIP
\Statex \textit{Step 2: query-relative screening (Eq.~\eqref{eq:m1})}
\State $P\gets\{n\in P_{\rm raw}:\ n\notin Q,\ \neg\mathcal R(q,n)\ \forall q\in Q\}$, no refill
\State keep the longest prefix of $P$ for which $Q\cup P$ fits the token budget
\Statex \textit{Step 3: condition-only readout (Eq.~\eqref{eq:interface})}
\State $(H,B)\gets D(x,\,Q\cup P)$ \Comment{one conditioned pass}
\State $s_{ic}\gets f_c(H_i,T_c)$ for every candidate $i$ and every $c\in Q$
\State \Return boxes $B$ with scores $s_{ic}$, $c\in Q$ only
\end{algorithmic}
\end{algorithm}

\begin{table*}[t]
\centering\small\setlength{\tabcolsep}{3pt}\renewcommand{\arraystretch}{1.15}
\caption{\textbf{Each step of NoM answers a measured finding.} All numbers are taken from the main paper or the sections cited. \mbox{GD-L} / \mbox{LLMDet-L}, Q8 unless stated.}
\label{tab:steps}
\begin{tabular}{p{0.11\textwidth}p{0.24\textwidth}p{0.18\textwidth}p{0.25\textwidth}p{0.14\textwidth}}\toprule
\raggedright Step & \raggedright Required by & \raggedright Fixed rule & \raggedright Evidence that it helps & \raggedright Learned alternative\tabularnewline\midrule
\raggedright \textbf{1. Scene nomination} & \raggedright Names of present objects carry most of the lost context. Annotated present names add 12.43 / 11.04 AP50, absent names 0.65 / 1.19, and neither can be output (Sec.~\ref{sec:measurement}, Table~\ref{tab:oracle}). & \raggedright A frozen CLIP image encoder ranks 1,203 LVIS names and keeps the top $k=8$, fixed on a separate 2,000-image set. & \raggedright Raw names cut TO8/100 from 52.09 to 10.44 and from 58.30 to 24.35, and exceed sinks, equal-count random names, and CLIP veto (Sec.~\ref{app:control_details}). Changing $k$ to 4 or 16, or the proposer to ViT-B/32, moves AP within intervals that include zero (Sec.~\ref{app:robustness}). & \raggedright Trained prompt pools stay below NoM in Q8 AP on both detectors (Sec.~\ref{app:additional_controls}).\tabularnewline
\raggedright \textbf{2. Query-relative screening} & \raggedright Direction rule (Eq.~\eqref{eq:rule}): a name moves a queried score with the sign of $g^\top\bar A_P-g^\top A_Q$, so a name close to the query can lower it (Sec.~\ref{sec:local}). & \raggedright Remove names related to any $q\in Q$ by WordNet synonymy, hypernymy, part/member/substance relations, or word containment (Eq.~\eqref{eq:m1}). & \raggedright Over the same raw list, AP rises by 0.51 [0.16, 0.96] and 1.74 [0.97, 2.51], with gains also at Q16 and full80 (Table~\ref{tab:mn}). & \raggedright Neither Presence nor Utility has a confirmed gain over NoM in any cell (Secs.~\ref{app:mn},~\ref{app:n2}).\tabularnewline
\raggedright \textbf{3. Condition-only readout} & \raggedright Label competition explains little of the loss. Open readout then filtering is only 0.35 / 0.21 AP50 below Q-only full conditioning (Sec.~\ref{sec:measurement}), and the user fixes the answer set. & \raggedright The detector reads $Q\cup P$, but scores are computed only against the query columns $T_Q$ (Eq.~\eqref{eq:interface}). & \raggedright Proposed names never become reports. Using CLIP at the output instead (CLIP veto) lowers Q8 AP by 3.79 / 2.91 (Table~\ref{tab:main}). & \raggedright No parameter to learn.\tabularnewline\bottomrule
\end{tabular}
\end{table*}

\noindent\textbf{When to apply NoM.} Across the eight text-conditioned checkpoints, the gain of NoM follows the native omission cost (Spearman 0.83, exact one-sided $p=0.0077$, Fig.~\ref{fig:cost_gain}). The omission cost needs only two native evaluations, so it indicates in advance whether NoM is worth applying (Sec.~\ref{sec:limits}). The same relation appears on ODinW-13 (Spearman 0.53, Sec.~\ref{app:odinw}). NoM is not needed where the omission cost is near zero, as for caption-derived queries (Sec.~\ref{app:further}).

\noindent\textbf{Alternative explanations.} Each competing explanation of the gain is tested by a control that changes one factor. Output label competition explains little: proposed names are never reported, withdrawal is measured with the readout fixed (Eq.~\eqref{eq:fixed}), and open readout followed by filtering is only 0.35 and 0.21 AP50 below Q-only readout under full conditioning (Sec.~\ref{sec:measurement}). Extra text tokens do not explain it either. Absent names add 0.65 and 1.19 AP50 against 12.43 and 11.04 for annotated present names (Table~\ref{tab:oracle}), and equal-count random names and 24 attention sinks stay below raw proposed names on both detectors (Sec.~\ref{app:control_details}). CLIP recognition used directly at the output (CLIP veto) lowers Q8 AP by 3.79 and 2.91 (Table~\ref{tab:main}), so the gain comes through conditioning. The concept filter is isolated by switching it on over the same proposal list (Table~\ref{tab:mn}). Tuning to one query set, proposer, or budget is checked with five query draws, four proposers, and $k\in\{4,8,16\}$ (Secs.~\ref{app:robustness} and~\ref{app:further}). Predictions about where NoM should and should not help were fixed before the runs, and those that failed are reported as failed (Secs.~\ref{app:a1_details}, \ref{app:odinw}, and~\ref{app:further}).

\begin{table}[t]
\centering\small\setlength{\tabcolsep}{3pt}
\caption{\textbf{Where each main-paper claim is supported.}}
\label{tab:claimmap}
\begin{tabular}{p{0.54\linewidth}p{0.11\linewidth}p{0.27\linewidth}}\toprule
\raggedright Claim & \raggedright Main & \raggedright Supplement\tabularnewline\midrule
\raggedright Partial queries cost accuracy and reorder detectors & \raggedright \ref{sec:measurement} & \raggedright \ref{app:a1_details}, \ref{app:expanded_details}\tabularnewline
\raggedright Names of present objects carry the lost context & \raggedright \ref{sec:measurement} & \raggedright \ref{app:visual_comparisons}\tabularnewline
\raggedright The cost enters at a detector-specific fusion site, and direction predicts the response & \raggedright \ref{sec:measurement}, \ref{sec:local} & \raggedright \ref{app:a1_details}, \ref{app:local}, \ref{app:score_accounting}, \ref{app:added_evidence}, \ref{app:further}\tabularnewline
\raggedright NoM recovers most of the cost, and each step contributes & \raggedright \ref{sec:main} & \raggedright \ref{app:a2_extension}, \ref{app:mn}, \ref{app:control_details}, \ref{app:added_evidence}, \ref{app:robustness}, \ref{app:further}\tabularnewline
\raggedright The gain follows the omission cost, and NoM is not needed when little is missing & \raggedright \ref{sec:limits} & \raggedright \ref{app:a4_details}, \ref{app:a3_extension}, \ref{app:optional_nom}, \ref{app:oracle_gate}, \ref{app:ow_details}, \ref{app:added_evidence}, \ref{app:robustness}, \ref{app:odinw}, \ref{app:further}\tabularnewline
\raggedright Learned alternatives do not improve on the fixed recipe & \raggedright \ref{sec:limits} & \raggedright \ref{app:additional_controls}, \ref{app:pool}, \ref{app:mn}, \ref{app:n2}\tabularnewline
\raggedright The two roles of the class list separate in a generative detector & \raggedright \ref{sec:generative} & \raggedright \ref{app:generative}\tabularnewline\bottomrule
\end{tabular}
\end{table}

\setcounter{table}{\value{savedtable}}\let\thetable\origthetable\let\theHtable\origtheHtable
\noindent\textbf{Reporting.} Every learned alternative was evaluated against success tests fixed before scoring, and every outcome is reported. None exceeds the fixed recipe consistently, which is why NoM remains training-free. NoM already comes close to annotated present names, within 0.69 AP on \mbox{GD-L} and 2.44 AP on \mbox{LLMDet-L} at Q8 (Tables~\ref{tab:main} and~\ref{tab:oracle}), which bounds what any better choice of names can add. Unless stated otherwise, intervals are paired 95\% image-bootstrap intervals computed for each contrast (marginal intervals). A difference is called confirmed when its interval excludes zero, and an interval that includes zero establishes neither a difference nor equivalence. The decoder gate in Sec.~\ref{app:oracle_gate} uses annotations and is a probe, not part of NoM. Each section below opens with the main-paper claim it supports (Table~\ref{tab:claimmap}).

\noindent\textbf{Reimplementation.} NoM has no trained parameters. It needs the public CLIP ViT-L/14 image encoder, the 1,203 LVIS category names, the WordNet relation of Eq.~\eqref{eq:m1}, $k=8$, the 256-token prefix rule, and the report rule of Sec.~\ref{sec:main} (900 candidates, one label each, at most 300 reports, score $>0.05$). Algorithm~\ref{alg:nom} lists the full procedure, and Sec.~\ref{app:a1_details} gives the evaluation configuration.

\noindent\textbf{Notation.} NoM always denotes the concept-filtered image-proposed names. The raw-name baseline is the same nomination without the concept filter. Box check adds detector-based presence confirmation. Presence and Utility are learned name scorers with different training targets (Secs.~\ref{app:mn} and~\ref{app:n2}). The decoder-gating probe of Sec.~\ref{app:oracle_gate} starts from NoM.

\noindent\textbf{Terms.} Q8 and Q16 query 8 and 16 classes per image, and full80 queries all 80. Native is the frozen detector prompted with the query alone. Full$\to$Q8 prompts with the full vocabulary and keeps only reports of the queried classes, and the omission cost is its AP50 minus native AP50 on the same images and queried classes. Sinks are 24 extra text tokens set to the mean of the frozen word-embedding table and never reported. TO8/100 counts unqueried GT taken over by queried-class reports, per 100 unqueried GT at Q8. A-OSE counts known-class reports matched to unknown objects in the open-world test. Both, or NoM + 24 sinks, adds the sinks to NoM. $\Delta$ is arm minus reference, in AP or AP50 points (pp).

\section{Raw-name gains across query counts}
\label{app:a4_details}
\supports{sec:limits}The benefit of added names shrinks as the query covers more classes and turns negative once all 80 names are given. Relative to native, \mbox{GD-L} gains $12.45$ AP50 $[9.60,13.75]$ at Q8, $8.58$ $[7.05,9.67]$ at Q16, and $2.42$ $[1.54,3.11]$ at Q40. \mbox{LLMDet-L} gains $6.49$ $[4.53,7.71]$, $3.33$ $[2.06,4.55]$, and $1.01$ $[0.36,1.70]$ at the corresponding query sizes. Full-context then filtering remains stronger than raw names at those settings. With all 80 names already supplied, the same fixed raw-name context changes \mbox{GD-L} AP50 by $-2.82$ $[-3.32,-2.35]$ and standard AP by $-2.38$ points. \mbox{LLMDet-L} AP50 changes by $-1.73$ $[-1.98,-1.39]$, and standard AP falls from $59.00$ to $57.21$. A frozen detector can still lose accuracy when context is added to a complete prompt.

\begin{table*}[tbp]\centering\small\setlength{\tabcolsep}{4pt}
\caption{\textbf{\mbox{LLMDet-L} open-world results.} Known AP50 is the authors' metric. COCO AP at IoU $0.50{:}0.95$ is not measured here. Known recall is the mean over known classes. Unknown AP50/recall are zero for every row.}
\label{tab:a3points}\begin{tabular}{rllrrr}\toprule Known & Score rule & Conditioning & Known AP50 & A-OSE & Known recall\\\midrule
20 & $>0.05$ & Native & 73.11 & 21,544 & 96.63\\
20 & $>0.05$ & Object & 73.74 & 24,617 & 96.62\\
20 & $>0.05$ & NoM & 69.48 & 17,654 & 95.65\\
20 & $>0.05$ & NoM + 24 sinks & 70.43 & 16,699 & 95.49\\
20 & No floor & Native & 73.12 & 41,630 & 96.78\\
20 & No floor & Object & 73.75 & 43,634 & 96.76\\
20 & No floor & NoM & 69.50 & 38,815 & 96.31\\
20 & No floor & NoM + 24 sinks & 70.44 & 40,295 & 96.13\\
40 & $>0.05$ & Native & 65.47 & 13,074 & 94.45\\
40 & $>0.05$ & Object & 65.98 & 14,460 & 94.20\\
40 & $>0.05$ & NoM & 63.60 & 11,348 & 93.75\\
40 & $>0.05$ & NoM + 24 sinks & 64.15 & 10,901 & 93.61\\
40 & No floor & Native & 65.48 & 26,507 & 94.62\\
40 & No floor & Object & 65.99 & 27,657 & 94.37\\
40 & No floor & NoM & 63.61 & 24,769 & 94.22\\
40 & No floor & NoM + 24 sinks & 64.18 & 25,460 & 94.17\\
\bottomrule\end{tabular}\end{table*}

\section{Open-world transfer with a restricted name bank (LLMDet-L)}
\label{app:a3_extension}
\supports{sec:limits}On \mbox{LLMDet-L}, NoM lowers open-set error at both vocabulary stages but also lowers known AP50, so the known-AP gain seen on \mbox{GD-L} (Sec.~\ref{app:ow_details}) does not carry over to this detector. The same detector shows no gain on the fixed LVIS query and loses on PascalVOC (Secs.~\ref{app:robustness} and~\ref{app:odinw}), so the weaker transfer appears specific to \mbox{LLMDet-L} and not to the open-world setting.
The evaluation uses the M-OWODB test list (10,246 rows, 9,119 distinct image files, 9,058 integer-ID clusters). All four arms share the frozen \mbox{LLMDet-L}, the authors' 100-report budget, the legal conditioning-name bank, and known-only output. Because only known classes are reported, unknown AP50 and recall are zero by design. The 20- and 40-class stages are vocabulary stages of one frozen detector, with no retraining between them.
Reports are kept at score $>0.05$, and a run without a score floor serves as a sensitivity analysis. Intervals use 1,000 paired draws of integer-ID clusters, keeping list multiplicities and a fixed tie order. Every point estimate matches a separate run of the official evaluator for all eight conditions in each stage. Tables~\ref{tab:a3points} and~\ref{tab:a3pairs} give absolute results and paired contrasts.
\begin{table*}[tbp]\centering\small\setlength{\tabcolsep}{4pt}
\caption{\textbf{All six primary-rule paired \mbox{LLMDet-L} contrasts at each stage.} Right arm minus reference. Marginal 95\% intervals and exact recovered/lost known-GT counts. Lower A-OSE is better.}
\label{tab:a3pairs}\begin{tabular}{rllrrr}\toprule Known & Arm & Reference & $\Delta$ known AP50 [95\% CI] & $\Delta$ A-OSE [95\% CI] & Recovered/lost\\\midrule
20 & Object & Native & +0.63 [0.37, 0.87] & +3,073.00 [2,775.00, 3,370.05] & 56/62\\
20 & NoM & Native & $-$3.63 [$-$4.27, $-$3.01] & $-$3,890.00 [$-$4,316.30, $-$3,457.93] & 106/217\\
20 & NoM & Object & $-$4.26 [$-$4.81, $-$3.68] & $-$6,963.00 [$-$7,429.27, $-$6,496.98] & 109/214\\
20 & NoM + 24 sinks & Native & $-$2.69 [$-$3.29, $-$2.05] & $-$4,845.00 [$-$5,306.20, $-$4,403.00] & 144/254\\
20 & NoM + 24 sinks & Object & $-$3.32 [$-$3.83, $-$2.74] & $-$7,918.00 [$-$8,454.02, $-$7,414.80] & 145/249\\
20 & NoM + 24 sinks & NoM & +0.95 [0.76, 1.18] & $-$955.00 [$-$1,180.03, $-$747.98] & 121/120\\
40 & Object & Native & +0.51 [0.31, 0.68] & +1,386.00 [1,152.92, 1,608.05] & 76/86\\
40 & NoM & Native & $-$1.87 [$-$2.31, $-$1.37] & $-$1,726.00 [$-$2,056.03, $-$1,399.00] & 175/334\\
40 & NoM & Object & $-$2.38 [$-$2.77, $-$1.92] & $-$3,112.00 [$-$3,450.25, $-$2,772.00] & 164/313\\
40 & NoM + 24 sinks & Native & $-$1.32 [$-$1.78, $-$0.78] & $-$2,173.00 [$-$2,523.07, $-$1,852.97] & 197/373\\
40 & NoM + 24 sinks & Object & $-$1.83 [$-$2.22, $-$1.33] & $-$3,559.00 [$-$3,948.00, $-$3,193.97] & 191/357\\
40 & NoM + 24 sinks & NoM & +0.55 [0.24, 0.82] & $-$447.00 [$-$591.00, $-$309.98] & 152/169\\
\bottomrule\end{tabular}\end{table*}
The stage-20 joint prediction, fixed before the run, is not met: NoM minus native is $-$3.63 [$-$4.27, $-$3.01] known AP50 and $-$3,890.00 [$-$4,316.30, $-$3,457.93] A-OSE. No prediction was fixed for stage 40. There, adding sinks to NoM lowers A-OSE under the $>0.05$ rule but changes it by $+691.00$ [473.95, 929.05] without a score floor, so the combination reduces A-OSE only under the primary rule. Nomination keeps the top eight names before the 256-token cap, with no refill.

\begin{table*}[tbp]\centering\small\setlength{\tabcolsep}{4pt}
\caption{\textbf{Additional error rows.} TO8/100 is taken-over GT per 100 unqueried GT. CC1/CC8 are percentages of images with a false report when querying 1/8 absent classes. Lower is better. Every row contains all four arms, including ties and reversals.}
\label{tab:a2error}\begin{tabular}{llrrrrl}\toprule Checkpoint & Error & Native & Sinks & NoM & Both & Strict order\\\midrule
\mbox{GLIP-TB} & TO8/100 & 60.17 & 59.17 & 56.37 & 55.56 & Pass\\
\mbox{GLIP-TB} & CC1 & 100.00 & 98.80 & 94.04 & 82.56 & Pass\\
\mbox{GLIP-TB} & CC8 & 100.00 & 100.00 & 99.88 & 99.68 & Fail\\
GD-Tiny & TO8/100 & 33.06 & 22.79 & 26.03 & 16.70 & Fail\\
GD-Tiny & CC1 & 95.02 & 26.06 & 50.22 & 22.62 & Fail\\
GD-Tiny & CC8 & 97.08 & 82.40 & 90.38 & 72.60 & Fail\\
\bottomrule\end{tabular}\end{table*}

\section{Error rates and AP on additional fusion checkpoints}
\label{app:a2_extension}
\supports{sec:limits}On \mbox{GLIP-TB} and GD-Tiny, adding names and sinks together gives the lowest point error rate in all six error rows. The stricter prediction, that every row follows native $>$ sinks $>$ names $>$ combination, holds in only two of the six rows, so the main-paper error ordering is stated for \mbox{GD-L} and \mbox{LLMDet-L} alone. Both checkpoints reuse the frozen NoM nominations, the 24 sinks, the Q8 query draw, the 5,000 images, the single-label output, and the score floor of the main two-detector table. GLIP keeps its dense candidate output, and GD-Tiny uses 900 decoder queries.

The order fails in two ways. On GLIP, CC8 is 100\% for both native and sinks, with difference interval $[0,0]$, so the first step is a tie at the prevalence ceiling. On GD-Tiny, sinks outperform names on all three error rates, which gives three further exceptions. Of the 18 adjacent intervals on these two checkpoints, 14 are negative, three positive, and one zero. Tables~\ref{tab:a2error} and~\ref{tab:a2adj} report error rates and adjacent differences, and Tables~\ref{tab:a2ap} and~\ref{tab:a2pairs} give AP and paired recognition changes.
\begin{table*}[tbp]\centering\small\setlength{\tabcolsep}{5pt}
\caption{\textbf{All adjacent error-rate contrasts.} Right minus left, marginal paired 95\% intervals from 5,000 image resamples. A zero interval is retained. The intervals do not provide simultaneous 95\% coverage.}
\label{tab:a2adj}\begin{tabular}{llrrr}\toprule Checkpoint & Error & Sinks$-$native & NoM$-$sinks & Both$-$NoM\\\midrule
\mbox{GLIP-TB} & TO8/100 & $-$1.00 [$-$1.31, $-$0.69] & $-$2.80 [$-$3.33, $-$2.27] & $-$0.82 [$-$1.09, $-$0.54]\\
\mbox{GLIP-TB} & CC1 & $-$1.20 [$-$1.50, $-$0.90] & $-$4.76 [$-$5.46, $-$4.10] & $-$11.48 [$-$12.40, $-$10.62]\\
\mbox{GLIP-TB} & CC8 & +0.00 [0.00, 0.00] & $-$0.12 [$-$0.22, $-$0.04] & $-$0.20 [$-$0.32, $-$0.08]\\
GD-Tiny & TO8/100 & $-$10.27 [$-$10.76, $-$9.78] & +3.24 [2.62, 3.89] & $-$9.34 [$-$9.79, $-$8.88]\\
GD-Tiny & CC1 & $-$68.96 [$-$70.24, $-$67.74] & +24.16 [22.76, 25.52] & $-$27.60 [$-$28.82, $-$26.34]\\
GD-Tiny & CC8 & $-$14.68 [$-$15.64, $-$13.70] & +7.98 [6.88, 9.08] & $-$17.78 [$-$18.80, $-$16.74]\\
\bottomrule\end{tabular}\end{table*}
\begin{table*}[tbp]\centering\small\setlength{\tabcolsep}{5pt}
\caption{\textbf{Official Q8 AP and object accounting for all four arms.} Each point is checked against a second official evaluator path. These absolute values apply to the pre-specified reporting adapter.}
\label{tab:a2ap}\begin{tabular}{llrrrr}\toprule Checkpoint & Arm & AP & AP50 & TP at IoU50 & FP at IoU50\\\midrule
\mbox{GLIP-TB} & Native & 23.41 & 30.04 & 3,363 & 1,486,191\\
\mbox{GLIP-TB} & Sinks & 23.68 & 30.22 & 3,373 & 1,477,971\\
\mbox{GLIP-TB} & NoM & 26.19 & 34.08 & 3,371 & 1,415,209\\
\mbox{GLIP-TB} & NoM + sinks & 26.16 & 34.04 & 3,383 & 1,365,312\\
GD-Tiny & Native & 46.13 & 59.49 & 3,634 & 135,433\\
GD-Tiny & Sinks & 48.88 & 63.49 & 3,589 & 74,886\\
GD-Tiny & NoM & 46.64 & 61.03 & 3,609 & 149,634\\
GD-Tiny & NoM + sinks & 49.81 & 65.17 & 3,602 & 77,065\\
\bottomrule\end{tabular}\end{table*}
\begin{table*}[tbp]\centering\small\setlength{\tabcolsep}{3pt}
\caption{\textbf{Q8 paired AP changes and queried-object recovered/lost counts.} AP improvement and error-rate ordering are evaluated separately.}
\label{tab:a2pairs}\begin{tabular}{llrrr}\toprule Checkpoint & Contrast & $\Delta$AP [95\% CI] & $\Delta$AP50 [95\% CI] & Recovered/lost\\\midrule
\mbox{GLIP-TB} & Sinks$-$native & +0.26 [$-$0.08, 1.17] & +0.18 [$-$0.27, 1.63] & 27/17\\
\mbox{GLIP-TB} & NoM$-$sinks & +2.51 [0.83, 3.87] & +3.86 [1.54, 5.47] & 41/43\\
\mbox{GLIP-TB} & Both$-$NoM & $-$0.03 [$-$0.74, 0.46] & $-$0.03 [$-$0.72, 0.56] & 19/7\\
\mbox{GLIP-TB} & NoM$-$native & +2.78 [1.44, 4.44] & +4.04 [2.46, 6.07] & 47/39\\
\mbox{GLIP-TB} & Both$-$native & +2.74 [1.53, 4.39] & +4.00 [2.57, 6.15] & 56/36\\
GD-Tiny & Sinks$-$native & +2.74 [2.20, 3.34] & +3.99 [3.25, 4.75] & 18/63\\
GD-Tiny & NoM$-$sinks & $-$2.24 [$-$3.71, $-$1.10] & $-$2.45 [$-$4.13, $-$1.14] & 68/48\\
GD-Tiny & Both$-$NoM & +3.17 [2.57, 4.00] & +4.13 [3.35, 5.02] & 27/34\\
GD-Tiny & NoM$-$native & +0.50 [$-$0.92, 1.60] & +1.54 [$-$0.11, 2.84] & 38/63\\
GD-Tiny & Both$-$native & +3.67 [2.42, 4.78] & +5.67 [4.11, 6.87] & 36/68\\
\bottomrule\end{tabular}\end{table*}

Near the prevalence ceiling, CC8 records whether an image receives any false report, not how many it receives. Q8 report totals are not false-positive counts either, and the official matched TP/FP in Table~\ref{tab:a2ap} give those. All 5,000 images enter the paired resampling, including the 171 images without an unqueried GT, and TO8 has denominator 32,464 for every checkpoint.

\begin{table*}[tbp]\centering\small\setlength{\tabcolsep}{4pt}

\caption{\textbf{Additional NoM Q8 tests.} 5,000 images and 5,000 paired image draws. Recovered/lost are IoU50 queried-GT object counts, independently verified from the official evaluator. These use checkpoint-specific reporting adapters, not the authors' leaderboard endpoints.}

\label{tab:optional_nom}\begin{tabular}{lrrrr}\toprule Checkpoint & Native AP50 & NoM AP50 & $\Delta$AP50 [95\% CI] & Recovered/lost\\\midrule
\mbox{GD-B} & 72.44 & 67.64 & $-$4.81 [$-$6.14, $-$3.47] & 22/316\\
\mbox{HDINO-T} & 66.78 & 64.67 & $-$2.11 [$-$2.72, 0.06] & 19/144\\
YOLO-World & 18.43 & 19.24 & +0.80 [$-$0.17, 1.25] & 30/29\\
\mbox{OV-DEIM-S} & 58.74 & 57.12 & $-$1.62 [$-$2.27, $-$0.69] & 35/83\\
\bottomrule\end{tabular}\end{table*}

\section{NoM on additional checkpoints}

\label{app:optional_nom}

\supports{sec:limits}The four further checkpoints have little or no omission cost, so NoM, applied without retuning, has little to recover on them. \mbox{GD-B} and \mbox{OV-DEIM-S} lose AP50, and \mbox{HDINO-T} and YOLO-World change within intervals that include zero (Table~\ref{tab:optional_nom}). Nomination, the concept filter, the Q8 query list, and the readout are those of the main paper, so proposed names never enter the output argmax.

Figure~\ref{fig:cost_gain} places these results at the low-cost end of the cost--gain trend, where NoM is not expected to help. The loss on \mbox{GD-B} does not come from extra output labels, because both arms report only the Q8 classes. Full80 NoM was not evaluated on these checkpoints.

\section{Decoder gating as an upper-bound probe}
\label{app:oracle_gate}
\supports{sec:limits}The gain of NoM is not confined to the decoder queries of unnamed objects. A gate that shows proposed names only to those queries keeps most of the gain, and so does its inverse. At full80, hiding the proposed names from the decoder leaves most of the cost that added names cause on a complete prompt.
The gate is chosen with annotations, so the probe is a diagnostic and not part of NoM. It runs on a fixed 300-image hash slice of the 2,000-image development set. Starting from NoM, each of the 900 final ungated boxes is assigned to a missing-name query when its IoU with an eligible present-but-unqueried GT exceeds $0.5$. A second pass shows the proposed-name tokens only to these queries (oracle), only to the remaining queries (inverse), or to a seeded random set of the same size. The gate acts in all six decoder text-attention layers and leaves the queried-name tokens and all computation before the decoder unchanged, so final boxes may still move (Table~\ref{tab:oracle_gate}).
\begin{table}[tbp]\centering\small\setlength{\tabcolsep}{3pt}
\caption{\textbf{Decoder-gating diagnostic, AP50.} 300 development images, point estimates only. Both vocabularies use the frozen concept-filtered NoM proposals. GT determines the oracle/inverse gates only in this diagnostic. No interval or significance claim is made. Differences use unrounded values.}
\label{tab:oracle_gate}\begin{tabular}{llrr}\toprule
Query & Condition & AP50 & $\Delta$ vs NoM\\\midrule
Q8 & Native &70.09&$-9.76$\\
 & NoM &79.85&0.00\\
 & Oracle gate &79.09&$-0.76$\\
 & Inverse gate &78.77&$-1.08$\\
 & Matched-count random &77.24&$-2.61$\\\midrule
Full80 & Native &69.79&$+2.02$\\
 & NoM &67.78&0.00\\
 & Decoder context off &68.11&$+0.33$\\\bottomrule
\end{tabular}\end{table}
The oracle gate falls $0.76$ AP50 below NoM, outside the $0.5$-point tolerance set in advance, while the inverse gate still gains $8.68$ over native. If the benefit sat only in the queries of missing objects, the oracle gate would match NoM and the inverse gate would stay close to native. Neither happens. The probe tests this gate and leaves other object-dependent ways of routing context open.

At full80 no present class is unqueried, so the oracle gate hides all proposed tokens from the decoder (Decoder context off in Table~\ref{tab:oracle_gate}). This recovers $0.33$ points but leaves a $1.69$-point cost relative to native, above the $0.5$-point target. Most of the cost of adding names to a complete prompt thus arises outside decoder text attention, in feature enhancement, text-conditioned query selection, query initialization, or text readout, and the probe does not separate these paths.

As a check on the hooks, an all-visible gate reproduces all 900 boxes, class scores, and hidden states exactly on three images at both query sizes, and all 300 ungated NoM reports match the reference.

\begin{table*}[tbp]\centering\small\setlength{\tabcolsep}{4pt}

\caption{\textbf{All pre-specified omission-cost contrasts.} AP50 points. Each cell uses the same 5,000-image paired resampling protocol. Negative values are retained. Rows follow the full80 order of Table~\ref{tab:rankscale}.}

\label{tab:a1all}\begin{tabular}{lrrr}\toprule Checkpoint & Q8 & Q16 & Q40\\\midrule
LLMDet & +12.93 [10.71, 14.22] & +9.22 [7.89, 10.14] & +4.32 [3.70, 4.81]\\
\mbox{GD-B} & +2.96 [1.94, 4.02] & +1.47 [0.54, 2.23] & +1.14 [0.71, 1.51]\\
WeDetect & +0.17 [$-$0.79, 0.56] & $-$0.33 [$-$0.59, $-$0.05] & $-$0.33 [$-$0.46, $-$0.22]\\
\mbox{GD-L} & +16.21 [13.59, 17.35] & +12.00 [10.53, 13.05] & +5.84 [5.19, 6.49]\\
GD-Tiny & +8.94 [7.32, 10.40] & +6.21 [5.07, 7.21] & +3.04 [2.44, 3.59]\\
\mbox{HDINO-T} & $-$1.16 [$-$1.78, 0.83] & +0.36 [$-$0.40, 1.05] & $-$0.13 [$-$0.55, 0.23]\\
\mbox{OV-DEIM-S} & $-$0.73 [$-$1.71, $-$0.11] & $-$0.48 [$-$1.07, $-$0.04] & $-$0.21 [$-$0.69, 0.12]\\
OWLv2 & $-$0.37 [$-$1.37, 0.42] & $-$0.26 [$-$0.85, 0.24] & $-$0.11 [$-$0.38, 0.14]\\
YOLOE & $-$0.17 [$-$0.75, 0.41] & $-$0.47 [$-$0.90, $-$0.08] & $-$0.31 [$-$0.53, $-$0.12]\\
\mbox{GLIP-TB} & +3.86 [2.40, 6.24] & +1.05 [0.26, 2.44] & +0.65 [$-$0.12, 1.46]\\
YOLO-World & +0.97 [0.12, 1.74] & +0.48 [$-$0.20, 0.91] & +0.28 [$-$0.01, 0.52]\\
\bottomrule\end{tabular}\end{table*}

\section{Additional checkpoint evaluation and secondary diagnostics}

\label{app:a1_details}

\supports{sec:measurement}The omission cost is positive in six of the eight text-conditioned checkpoints and stays within one point in the three text-independent ones (Table~\ref{tab:a1all}). This section also gives the full configuration and secondary diagnostics.

\paragraph{Experimental configuration and reporting.}
The main benchmark uses all 5,000 COCO val2017 images~\cite{coco}. Q8, Q16, and full80 query 8, 16, and all 80 classes. The query-count analysis also includes Q40. Q8 is primary, and full80 tests complete prompts. COCO validation and the M-OWODB/S-OWODB test sets were also used during development, and the NoM configuration was fixed on a separate 2,000-image selection set.

Grounding DINO-family comparisons use 900 candidates, one label per candidate, at most 300 reports per image, and score $>0.05$. Fixed image-hash queries are independent of annotations. AP is averaged over IoU thresholds $0.50{:}0.05{:}0.95$. AP50 uses IoU $0.50$. Timing includes nomination, with scopes stated locally. NoM's selection rules are in Sec.~\ref{sec:nom}.

\paragraph{Recognition and error counts.}
Takeover is the number of unqueried GT covered by queried-class reports, per 100 eligible unqueried GT. Commission cost is the percentage of eligible images with any false report when all queried classes are absent (Sec.~\ref{app:visual_comparisons}). Neither count determines AP, which also depends on true positives and ranking.

Unless noted, intervals use 5,000 paired image resamples: all objects from one image share a resampling unit. Checkpoints evaluated on the same images are not independent physical samples. Controls are equal-count random names, four fixed names plus sinks, and CLIP veto at the fixed threshold $0.5$ (Sec.~\ref{app:expanded_details}).

The scale experiment retains each checkpoint's pre-specified candidate generation and published single-label endpoint. The comparison does not equalize compute or training data. GLIP and YOLO-World use dense adapters, not the authors' multilabel/NMS evaluation. The low YOLO-World absolute AP applies to that adapter only. \mbox{GD-B} is COCO-supervised. OWLv2 uses the official square-padding geometry. YOLOE uses its text-mode checkpoint and class-aware NMS. WeDetect uses 33,600 dense candidates, pre-NMS 30,000, NMS 0.7, and the authors' Chinese text bank, whereas GD-family models use 900 decoder queries and BERT text. Tables~\ref{tab:a1all} and~\ref{tab:family_means} give paired costs and family means.

The pre-specified broad text-conditioned group includes both feature fusion and text-dependent query selection. \mbox{OV-DEIM} is selection-only, so it is not assigned a decoder text cross-attention mechanism.

\begin{table}[tbp]\centering\small\setlength{\tabcolsep}{3pt}

\caption{\textbf{Descriptive family means.} AP50 points. Membership was fixed before results were opened.}
\label{tab:family_means}

\begin{tabular}{lrrr}\toprule Group & Q8 & Q16 & Q40\\\midrule

Conditioned (8) & +5.50 & +3.79 & +1.87\\

Independent (3) & $-$0.12 & $-$0.35 & $-$0.25\\

\bottomrule\end{tabular}\end{table}

These are descriptive arithmetic means over the pre-specified checkpoints, not population estimates over detector architectures. Three predictions were specified before this evaluation. First, all eight text-conditioned checkpoints would show a Q8 omission cost with a paired lower bound above zero. This is not met: six do, and \mbox{HDINO-T} and \mbox{OV-DEIM-S} do not. Second, all three text-independent checkpoints would show a point omission cost of at most +1 AP50. This is met, with $-$0.37 (OWLv2), $-$0.17 (YOLOE), and +0.17 (WeDetect). Third, a cross-group pair other than LLMDet and WeDetect would reverse its AP50 order between full80 and Q8. This is met at the point level by \mbox{GD-B} and WeDetect, whose Q8 interval includes zero (Sec.~\ref{app:expanded_details}).

\begin{figure}[tbp]
\centering\includegraphics[width=\linewidth]{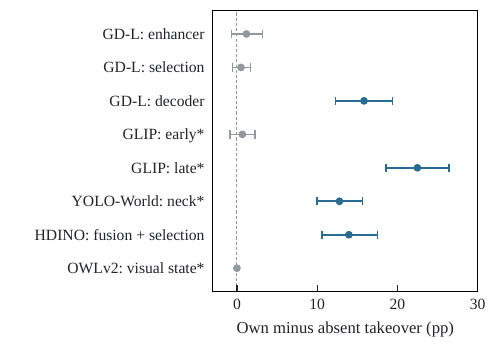}
\caption{\textbf{Where the controlled effect enters.} Earlier fixed-readout intervention on the 400-image development set (dev400), top 20 reports and no score floor. Architecture groups are separate interventions, not additive fractions. Asterisks mark interventions with fixed boxes. Other geometry can respond. These values do not use the COCO top-300 AP protocol of the main results.}
\label{fig:mechanism}
\end{figure}

In an earlier 1,000-image LVIS block-20 diagnostic, 77,291/78,189 short-prompt cross-block takeovers (98.85\%) involve classes without positive or verified negative annotations. The federated evaluator ignores them~\cite{lvis}. They are unverified reports, not certified errors or a detector-wide error rate.

\subsection{Additional analyses}
Conditioning accounts for 16.35 of the 27.94-point \mbox{GD-L} withdrawal effect and for none of the OWLv2 effect (Table~\ref{tab:paths}). The path comparison uses 400 development images and 587 targets, top 20 without a score floor, and 5,000 paired image draws (seed 20261109). Table~\ref{tab:paths} gives total, fixed-readout internal, and residual effects. OWLv2 fixed-readout reports are bitwise unchanged on all 400 images. These operational residuals are not natural causal fractions or substitutes for Fig.~\ref{fig:mechanism}.

\begin{table}[tbp]
\centering\small\setlength{\tabcolsep}{4pt}
\caption{\textbf{Matched path decomposition.} Points (pp), earlier dev400 \mbox{top-20}/no-floor endpoint. Residual is computed per paired draw from total minus fixed-readout internal effect. It is not an identified natural direct effect.}
\label{tab:paths}
\begin{tabular}{lrrr}\toprule
Model & Total & Internal & Residual\\\midrule
\mbox{GD-L} &27.94&16.35&11.58\\
\mbox{GD-B} &34.92&12.78&22.15\\
OWLv2 &24.70&0.00&24.70\\\bottomrule
\end{tabular}
\end{table}

A lexical check finds potentially related nominations in 49 of 80 classes, covering 1,080 of 40,000 Q8 query opportunities (2.70\%). Dictionary relations to the query are not limited to the train example of Fig.~\ref{fig:class_differences}. Many of these WordNet pairings are sense-ambiguous. Whether related names lower queried scores is tested directly in Sec.~\ref{app:further}: names removed by the concept rule lower the scores of true queried candidates more than retained names do, on both detectors (Table~\ref{tab:fu_insert}). 

The stored absent-query comparisons (native, earlier name-only context, sinks, combination) give Q8 false positives/image of 31.82/3.47/7.77/1.61 (\mbox{GD-L}) and 33.69/12.97/19.66/9.51 (\mbox{LLMDet-L}). Names and sinks each lower false positives, the combination lowers them most, and none removes them. These values use our reporting protocol, not a Tractor-Dogs reproduction.

\section{Matched learned alternatives to the fixed recipe}
\label{app:additional_controls}
\supports{sec:limits}Learning the context does not beat nominating it. Conditioning tokens trained on the same data with the detector frozen stay below NoM in Q8 AP on both detectors, for every pool (Tables~\ref{tab:trained_detailed} and~\ref{tab:main}), and the projected pools do not improve on the plain pool. Their advantage appears at full80, where the dual-projection pool stays within 0.2 AP50 of native, a setting in which NoM is not needed (Sec.~\ref{sec:limits}).
\subsection{Limits of projected prompt pools}
Each pool has 16 learned conditioning tokens and comes in four variants: plain, value-projected, key-projected, and dual-projected. All variants share the data, optimization, and frozen detector, and only the pool is trained (Fig.~\ref{fig:pool}, Sec.~\ref{app:pool}). Table~\ref{tab:trained_detailed} reports the matched outcomes.

The projections aim to let the pool add context without moving queried scores, but they constrain only local components. Value projection leaves the attenuation term $-m_PA_Q$ of Eq.~\eqref{eq:two} in place and can remove useful evidence, and key projection zeros inserted logits only within the queried-key span, so it does not bound attention mass in general (Sec.~\ref{app:pool}).

\begin{table*}[tbp]
\centering\small\setlength{\tabcolsep}{3pt}
\caption{\textbf{Matched learned controls at the 8,500-microstep / 1,062-update snapshots.} NoM reaches 52.18 (\mbox{GD-L}) and 56.94 (\mbox{LLMDet-L}) Q8 AP without training (Table~\ref{tab:main}).}
\label{tab:trained_detailed}
\begin{tabular}{llccccc}\toprule
Model & Method & Q8 AP & Q8 AP50 & Q16 AP50 & full80 AP & $t_{\rm Q8}$ (ms)\\
\midrule
\mbox{GD-L} & Ordinary local fine-tune &50.73&65.84&63.66&45.50&556.56\textsuperscript{$E$}\\
 & Plain pool &50.83&66.45&64.94&51.62&571.53\textsuperscript{$E$}\\
 & Value-projected pool &50.69&66.09&64.95&52.28&657.58\textsuperscript{$E$}\\
 & Dual-projection pool &48.79&63.31&63.81&52.61&691.94\textsuperscript{$E$}\\
\midrule
\mbox{LLMDet-L} & Ordinary local fine-tune &53.94&69.43&68.01&50.54&571.27\textsuperscript{$E$}\\
 & Plain pool &55.84&69.74&69.71&59.01&569.50\textsuperscript{$E$}\\
 & Value-projected pool &55.77&69.53&69.79&59.08&659.39\textsuperscript{$E$}\\
 & Dual-projection pool &54.05&67.34&67.45&58.95&706.10\textsuperscript{$E$}\\\bottomrule
\end{tabular}
\end{table*}
At the fixed snapshot, the dual-projection pool falls below the plain pool at Q8, with negative AP and AP50 intervals on both models, and the value-projected pool shows no confirmed gain over it. The local constraints do not deliver the intended short-query benefit, and the AP comparison does not show which evidence they remove. The Q8 tests of raw names against the plain pool and against local fine-tuning are not met either.

At full80 the dual-projection pool helps on \mbox{GD-L}. It exceeds the plain pool by $+0.99$ $[0.79,1.15]$ AP and $+1.14$ $[0.84,1.32]$ AP50, and local fine-tuning by $+7.11$ $[6.41,7.80]$ / $+8.91$ $[8.06,9.74]$, with recovered/lost GT of 593/240 and 2,856/836. Its AP50 of 69.09 is within the $0.5$-point tolerance of the same-set native 69.06. Against raw names it gains $+2.38$ AP (point value only) and $+2.85$ $[2.39,3.31]$ AP50. On \mbox{GD-L} the pool helps where NoM is not needed and not where NoM helps.

At Q16 the dual-projection pool changes AP50 relative to the plain pool by $-1.13$ $[-1.55,-0.72]$ on \mbox{GD-L} and $-2.26$ $[-2.81,-1.70]$ on \mbox{LLMDet-L}, with recovered/lost GT of 67/62 and 27/46, and both standard-AP intervals are negative. Against local fine-tuning it changes AP50 by $+0.15$ $[-1.12,1.43]$ and $-0.56$ $[-1.91,0.71]$, with recovered/lost GT of 386/186 and 467/38, and no AP or AP50 gain is confirmed.

On \mbox{LLMDet-L} at full80, the dual-projection pool changes AP by $-0.06$ $[-0.28,0.09]$ and AP50 by $-0.25$ $[-0.45,-0.09]$ relative to the plain pool, with 175 recovered and 305 lost GT. It exceeds local fine-tuning by $+8.41$ $[7.82,8.99]$ AP and $+9.06$ $[8.31,9.76]$ AP50, with 3,052/281 recovered/lost GT. Its AP50 of 75.83 against native 76.01, a paired change of $-0.18$ $[-0.28,-0.09]$, is also within the tolerance. Against raw names it gains $+1.74$ AP (point value only) and $+1.55$ $[1.19,1.80]$ AP50.

\paragraph{Learned name selection.}
Learned name selectors do not beat the fixed concept rule either. Box check confirms proposed names with detector evidence before a query-only second decode, Presence learns a query-independent presence score, and Utility predicts a name's effect under the current query. Box check and Presence pass their \mbox{LLMDet-L} Q8 tests against raw names, but no AP or AP50 gain of Presence over NoM is confirmed in any cell. Utility does not meet its Q8 test over Presence and has no confirmed gain over Presence or NoM. Sections~\ref{app:mn} and~\ref{app:n2} give the methods, tests, contrasts, and object records.

\section{Specification of the learned pools}
\label{app:pool}
\begin{figure}[tbp]
\centering\includegraphics[width=\linewidth]{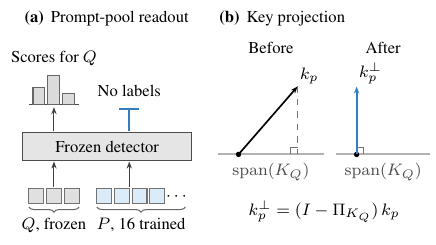}
\caption{\textbf{Learned-pool control (Sec.~\ref{app:additional_controls}).} (a) Sixteen trained prompt tokens $P$ enter the frozen detector with the query $Q$, but only $Q$ is scored as labels. (b) Each prompt key is replaced by its component orthogonal to the span of the queried keys. Prompt values are projected off the queried readout directions in the same way (Eq.~\eqref{eq:projections}). Only the pool is trained, and neither projection guarantees that the full detector does not degrade.}
\label{fig:pool}
\end{figure}

\supports{sec:local}This section specifies the learned pools of Sec.~\ref{app:additional_controls}, which are controls and not part of NoM (Fig.~\ref{fig:pool}). Their projections remove only local components of the update in Eq.~\eqref{eq:two}, so they cannot guarantee that queried objects are protected, consistent with the results there.
\subsection{Scene prompt pool with two projections}
The trainable extension uses 16 learned tokens of width 256, initialized from 16 k-means centers of fusion-input text features for the 1,203 LVIS names. The tokens participate in the encoder fusion and all six decoder text-attention layers, and never become output classes. Image conditioning can arise in the existing image-to-text fusion computation. No additional retrieval module is assumed. The detector weights are frozen, and only the pool is optimized.

For each attention layer, let $T_Q$ span the queried readout directions and $K_Q$ span their layer-specific projected keys. With orthogonal projectors $\Pi_{T_Q}$ and $\Pi_{K_Q}$, the two operations are
\begin{equation}
u_p^\perp=(I-\Pi_{T_Q})u_p,\qquad
k_p^\perp=(I-\Pi_{K_Q})k_p.
\label{eq:projections}
\end{equation}
The value operation makes $t_c^\top u_p^\perp=0$ for the specified queried directions. It removes the inserted-value term's direct linear projection. The dilution term $-m_PA_Q$ remains. Thus the derivative of the entire output along a queried direction is generally $-t_c^\top A_Q$, not zero. Moreover, a downstream gradient need not lie in the projected readout span. The value constraint gives neither whole-network score invariance nor guaranteed protection of true objects.

The key operation gives $q^\top k_p^\perp=0$ only when $q\in\operatorname{span}(K_Q)$. Under this additional condition, with no unaccounted logit bias or mask, every inserted logit is zero and $m_P=N/(Z_Q+N)$. This is a conditional identity, not a generally small upper bound: $Z_Q$ need not be large, and a real decoder query can have a component outside the queried-key span. Orthogonal keys thus need not prevent dilution, and the ablations measure attention mass and detector scores directly.

\subsection{Training and matched arms}
Each training query is a subset of the 80 COCO names, absent classes included, whose size is drawn uniformly from 1 to 16. A frozen full-80-name teacher and the student share the cached image backbone computation. Their top 300 queries are matched by a Hungarian assignment with L1 and generalized-IoU box costs. The objective combines queried-column binary cross-entropy distillation with a supervised queried-class detection loss,
\begin{equation}
\mathcal L=\mathcal L_{\rm distill,Q}+0.5\mathcal L_{\rm det,Q}.
\label{eq:loss}
\end{equation}
The supervised term uses only ground-truth classes in $Q$. The overall recipe is not annotation-free. The optimizer is AdamW, learning rate $10^{-3}$ with cosine decay. All arms use identical data and microsteps. The time-limited recipe uses a fixed hashed 30,000-image subset.

The plain pool has no projection. Value-only, key-only, and dual variants apply the corresponding projections. Local fine-tuning updates the detector without a pool and removes missing-name boxes from supervision. Raw names provides the training-free reference. Its projected variant applies both constraints to the supplied context. Table~\ref{tab:trained_detailed} evaluates 8,500 microsteps / 1,062 updates. A later set targeting 12,000 / 1,500 is a separate run and is not combined with these results. The principal comparison uses the snapshot fixed before scoring, not the best checkpoint after seeing validation scores.

\section{Additional local response quantities}\label{app:local}
\supports{sec:local}A susceptibility measure that needs no missing name was also tested. Unlike the signed direction predictor of Eq.~\eqref{eq:rule}, it does not meet its pre-specified test, so a simpler variance statistic does not replace the direction rule.
\subsection{Susceptibility without the missing name}
For a current query $i$, let $g_i$ be the gradient of its highest queried-class score with respect to the final decoder text-attention output. We use the pre-specified quantity
\begin{equation}
\chi_i=\Var_{j\sim p_i}\!\left[\langle g_i,W_ov_j\rangle\right].
\label{eq:chi}
\end{equation}
Its local interpretation follows by setting $r_j=\langle g_i,u_j\rangle$ and considering the specified exponential tilt $p_j(\epsilon)\propto p_j\exp(\epsilon r_j)$. For $\psi(\epsilon)=\log\sum_jp_j\exp(\epsilon r_j)$, $\psi''(0)=\Var_p(r)=\chi_i$. Equivalently, the derivative at zero of the tilted mean of $r$ is its variance. This is the log-partition curvature familiar from exponential-family information geometry~\cite{amari}. It is not an identity for the second derivative of an arbitrary nonlinear detector score. Gradients and values are frozen in this interpretation.

The implementation sums $\Var_{p_{ih}}[\langle g_i,W_{o,h}v_{hj}\rangle]$ across heads, using one local-prompt forward and backward pass. The pre-specified joint test requires the takeover-correlation lower bound to exceed $0.3$ and the confidence-controlled partial-correlation bound to exceed zero in each model. On the 400-image development set, $\chi$ does not meet this joint test. Its correlation with the score improvement from raw names is descriptive.

\section{Concept filter: full results}
\label{app:mn}
\supports{sec:main}Screening query relatives raises standard AP over the same raw list on both detectors at Q8, Q16, and full80, with no refill (Table~\ref{tab:mn}). It restores queried reports while takeover rates change little (Table~\ref{tab:takeover_rates}). Table~\ref{tab:m1wedetect} gives the cross-detector reference. Before filtering, the \mbox{LLMDet-L} raw-name AP50 gap to WeDetect at Q8 is 2.98 [0.99, 4.76]. WeDetect uses 33,600 dense candidates and the authors' Chinese text bank, versus 900 decoder queries in the GD family (Sec.~\ref{app:a1_details}).
All rows use the 5,000 images, the same per-image queries, 900 candidates, one output label per candidate, top 300 reports, and score $>0.05$. Paired intervals use 5,000 image resamples with seed 20261110. NoM filters only the eight raw names, with no replacement. Box check uses threshold $0.3$ fixed on the 2,000-image selection set and $k=8$. Its \mbox{LLMDet-L} Q8 test passes, while the negative \mbox{GD-L} short-query cells remain. Presence is evaluated below. Utility is reported separately in Sec.~\ref{app:n2}. Intervals characterize paired image variation, not training-seed variation. No multiple-comparison correction is applied.
\begin{table*}[tbp]
\centering\small\setlength{\tabcolsep}{3pt}
\caption{\textbf{NoM and Box check against raw names.} Deltas are relative to the same raw-name list. Intervals are paired 95\%. R/L are recovered/lost queried GT. A/N are avoided/new takeovers of unqueried GT. At full80 there are no unqueried GT, so takeover rate and A/N are inapplicable. These counts are not an AP decomposition.}
\label{tab:mn}
\begin{tabular}{lllrrllll}\toprule
Model&Arm&Query&AP&AP50&$\Delta$AP [95\% CI]&$\Delta$AP50 [95\% CI]&R/L&A/N\\\midrule
\mbox{GD-L}&NoM&Q8&52.18&68.58&+0.51 [0.16, 0.96]&+0.36 [$-$0.06, 0.83]&7/4&83/134\\
\mbox{GD-L}&NoM&Q16&51.15&66.94&+0.82 [0.39, 1.29]&+0.75 [0.26, 1.26]&24/11&126/290\\
\mbox{GD-L}&NoM&full80&50.73&66.73&+0.50 [0.33, 0.69]&+0.49 [0.28, 0.74]&236/140&--\\
\mbox{LLMDet-L}&NoM&Q8&56.94&72.15&+1.74 [0.97, 2.51]&+2.09 [1.21, 2.82]&9/1&110/266\\
\mbox{LLMDet-L}&NoM&Q16&55.24&70.63&+0.96 [0.60, 1.43]&+1.25 [0.79, 1.74]&17/6&193/361\\
\mbox{LLMDet-L}&NoM&full80&58.00&75.10&+0.79 [0.65, 0.92]&+0.82 [0.66, 0.97]&170/106&--\\
\mbox{GD-L}&Box check&Q8&51.16&66.06&$-$0.51 [$-$1.52, 0.53]&$-$2.16 [$-$3.37, $-$0.84]&97/66&706/2,804\\
\mbox{GD-L}&Box check&Q16&49.70&64.28&$-$0.63 [$-$1.45, 0.21]&$-$1.90 [$-$2.79, $-$0.77]&189/120&861/3,180\\
\mbox{GD-L}&Box check&full80&51.53&67.80&+1.31 [1.01, 1.65]&+1.56 [1.20, 2.03]&805/335&--\\
\mbox{LLMDet-L}&Box check&Q8&57.53&72.76&+2.33 [1.32, 3.27]&+2.69 [1.48, 3.63]&30/24&2,030/1,853\\
\mbox{LLMDet-L}&Box check&Q16&56.21&71.80&+1.92 [1.05, 2.56]&+2.41 [1.36, 3.02]&56/49&1,904/1,652\\
\mbox{LLMDet-L}&Box check&full80&58.33&75.29&+1.13 [0.84, 1.34]&+1.01 [0.70, 1.28]&275/260&--\\
\bottomrule\end{tabular}
\end{table*}

At Q8 the denominator is 32,464 valid, non-crowd GT whose classes are outside the image's query. The \mbox{GD-L} raw-name baseline takes over 3,388 of them, versus 3,439 for NoM and 5,486 for Box check. The \mbox{LLMDet-L} raw-name baseline takes over 7,904, versus 8,060 for NoM and 7,727 for Box check. Table~\ref{tab:takeover_rates} gives the rates with marginal intervals.
\begin{table}[tbp]
\centering\small\setlength{\tabcolsep}{4pt}
\caption{\textbf{Q8 takeover rates per 100 unqueried GT.} These marginal intervals are not confidence intervals for a difference.}
\label{tab:takeover_rates}
\begin{tabular}{llrl}\toprule
Model&Arm&Count&Rate [95\% CI]\\\midrule
\mbox{GD-L}&raw names&3,388&10.44 [9.84, 11.07]\\
\mbox{GD-L}&NoM&3,439&10.59 [10.00, 11.22]\\
\mbox{GD-L}&Box check&5,486&16.90 [16.10, 17.72]\\
\mbox{LLMDet-L}&raw names&7,904&24.35 [23.47, 25.22]\\
\mbox{LLMDet-L}&NoM&8,060&24.83 [23.94, 25.73]\\
\mbox{LLMDet-L}&Box check&7,727&23.80 [22.89, 24.71]\\\bottomrule
\end{tabular}
\end{table}

Q16 has 28,712 eligible unqueried GT. \mbox{GD-L} raw names/NoM/Box check takeover counts (rates per 100 with marginal 95\% intervals) are 4,773 (16.62 [15.85, 17.45]), 4,937 (17.19 [16.40, 18.04]), and 7,092 (24.70 [23.69, 25.79]). \mbox{LLMDet-L} raw names/NoM/Box check give 9,433 (32.85 [31.86, 33.90]), 9,601 (33.44 [32.42, 34.50]), and 9,181 (31.98 [30.96, 33.00]).

\begin{table}[tbp]
\centering\small\setlength{\tabcolsep}{3pt}
\caption{\textbf{NoM minus WeDetect, AP50 (pp).} 5,000 paired image draws. Cross-backbone candidate and compute budgets differ. Q8/full80 use different GT sets. Q16 and object counts were not computed. The standard-AP comparison is in Sec.~\ref{app:added_evidence}.}
\label{tab:m1wedetect}
\begin{tabular}{llrl}\toprule
Model&Query&$\Delta$AP50&Paired 95\% CI\\\midrule
\mbox{GD-L}&Q8&$-$4.47&[$-$6.51, $-$2.70]\\
\mbox{GD-L}&full80&$-$4.18&[$-$4.89, $-$3.33]\\
\mbox{LLMDet-L}&Q8&$-$0.90&[$-$2.61, 0.97]\\
\mbox{LLMDet-L}&full80&+4.19&[3.50, 4.86]\\\bottomrule
\end{tabular}
\end{table}

An earlier timing covered image-feature extraction plus conditioned decoding only (\mbox{GD-L} 209.91+366.39 ms, \mbox{LLMDet-L} 209.53+367.06 ms). The later end-to-end measurements in Table~\ref{tab:main}, 600.5 and 579.5 ms/image, include nomination and replace that timing (Fig.~\ref{fig:latency_plot}). Box check \mbox{GD-L} Q8 measures CLIP 21.43 ms, image features 214.84 ms, first conditioned decode 423.26 ms, and second decode 366.46 ms: a shared-backbone component sum of 1,026.00 ms. Charging the image pass twice gives 1,240.84 ms. These stage sums exclude loading and text-bank setup and are not a co-resident end-to-end FPS measurement.

\subsection{Presence confirmation as an alternative selector}
Presence, a learned presence confirmer specified in Sec.~\ref{app:n2}, passes its \mbox{LLMDet-L} Q8 test against raw names, but no gain over NoM is confirmed in any cell. Table~\ref{tab:nabsolute} gives the six Presence endpoints, and Table~\ref{tab:ncontrasts} gives all four contrasts in every cell. The comparison with NoM was added after the tests were fixed and is descriptive. At \mbox{LLMDet-L} Q8, the intervals of Presence against NoM, Box check, and the 32-proposal rule cross zero for both AP and AP50.

On \mbox{GD-L}, Presence exceeds the 32-proposal rule in Q8 and Q16 AP50 and in Q16 AP, but at neither query size is a gain over raw names or NoM confirmed in both metrics. Its \mbox{GD-L} full80 AP and AP50 are lower than those of Box check and the 32-proposal rule. Across all six cells, no AP or AP50 interval establishes Presence above NoM. \mbox{GD-L} Q16 AP50 differs from NoM by $-0.95$ $[-1.72,-0.03]$. Its \mbox{LLMDet-L} full80 AP is lower than that of Box check. These results concern this presence target, feature set, selection rule, and budget.

Object counts do not establish an independent AP gain: Presence versus NoM at \mbox{LLMDet-L} Q8 recovers 21 queried GT and loses 23, while avoiding 1,925 takeovers and introducing 1,493. Full80 has no unqueried GT. Image intervals exclude training-seed variability.

\begin{table*}[tbp]
\centering\small\setlength{\tabcolsep}{3pt}
\caption{\textbf{Presence absolute results on the 5,000 COCO validation images.} Intervals are paired-image marginal 95\% intervals. Matched/total concerns queried GT. TO counts taken-over unqueried GT. Full80 has no unqueried GT, so its takeover rate is not applicable (--).}
\label{tab:nabsolute}
\begin{tabular}{llllrrl}\toprule
Model&Query&AP [95\% CI]&AP50 [95\% CI]&Matched/total&TO&TO/100 [95\% CI]\\\midrule
\mbox{GD-L}&Q8&52.13 [50.99, 55.46]&67.90 [66.19, 71.22]&3,518/3,871&4,188&12.90 [12.21, 13.62]\\
\mbox{GD-L}&Q16&50.68 [50.24, 53.98]&65.99 [65.22, 69.36]&6,914/7,623&5,631&19.61 [18.73, 20.53]\\
\mbox{GD-L}&full80&50.69 [50.18, 52.02]&66.84 [66.09, 68.23]&32,331/36,335&0&--\\
\mbox{LLMDet-L}&Q8&57.21 [56.30, 60.46]&72.38 [71.01, 75.51]&3,718/3,871&7,628&23.50 [22.63, 24.37]\\
\mbox{LLMDet-L}&Q16&55.60 [55.03, 58.86]&71.25 [70.37, 74.47]&7,338/7,623&9,207&32.07 [31.06, 33.09]\\
\mbox{LLMDet-L}&full80&57.92 [57.37, 59.16]&75.11 [74.30, 76.27]&34,315/36,335&0&--\\
\bottomrule\end{tabular}
\end{table*}

\begin{table*}[tbp]
\centering\small\setlength{\tabcolsep}{3pt}
\caption{\textbf{Presence minus each of four references in all six cells (pp).} AP and AP50 use 5,000 paired image resamples, seed 20261110. NoM contrasts are additional and descriptive. The 32-proposal rule uses the same input evidence with threshold $0.2$ fixed on the 2,000-image selection set and $k=4$. R/L counts recovered/lost queried GT. A/N counts avoided/new takeovers. A/N is not applicable at full80 (--). No multiple-comparison correction is applied.}
\label{tab:ncontrasts}
\begin{tabular}{lll ll rr}\toprule
Model&Query&Reference&$\Delta$AP [95\% CI]&$\Delta$AP50 [95\% CI]&R/L&A/N\\\midrule
\mbox{GD-L}&Q8&raw names&+0.46 [$-$0.40, 1.41]&$-$0.33 [$-$1.37, 0.83]&86/50&743/1,543\\
\mbox{GD-L}&Q8&NoM&$-$0.05 [$-$0.85, 0.79]&$-$0.68 [$-$1.62, 0.39]&81/48&752/1,501\\
\mbox{GD-L}&Q8&Box check&+0.97 [0.13, 1.80]&+1.84 [0.79, 2.84]&52/47&1,987/689\\
\mbox{GD-L}&Q8&32-proposal rule&+0.82 [$-$0.01, 1.44]&+1.13 [0.16, 1.87]&39/36&1,180/711\\
\midrule
\mbox{GD-L}&Q16&raw names&+0.35 [$-$0.43, 1.14]&$-$0.20 [$-$1.04, 0.74]&135/99&911/1,769\\
\mbox{GD-L}&Q16&NoM&$-$0.47 [$-$1.14, 0.26]&$-$0.95 [$-$1.72, $-$0.03]&125/102&951/1,645\\
\mbox{GD-L}&Q16&Box check&+0.98 [0.41, 1.53]&+1.70 [0.92, 2.30]&90/123&2,197/736\\
\mbox{GD-L}&Q16&32-proposal rule&+1.01 [0.52, 1.54]&+1.19 [0.54, 1.76]&77/75&1,253/767\\
\midrule
\mbox{GD-L}&full80&raw names&+0.47 [0.14, 0.82]&+0.61 [0.20, 1.06]&549/360&--\\
\mbox{GD-L}&full80&NoM&$-$0.03 [$-$0.35, 0.29]&+0.12 [$-$0.29, 0.54]&472/379&--\\
\mbox{GD-L}&full80&Box check&$-$0.84 [$-$1.11, $-$0.57]&$-$0.95 [$-$1.33, $-$0.60]&271/552&--\\
\mbox{GD-L}&full80&32-proposal rule&$-$0.31 [$-$0.51, $-$0.06]&$-$0.38 [$-$0.62, $-$0.05]&256/339&--\\
\midrule
\mbox{LLMDet-L}&Q8&raw names&+2.01 [1.06, 2.97]&+2.31 [1.16, 3.29]&25/19&1,915/1,639\\
\mbox{LLMDet-L}&Q8&NoM&+0.27 [$-$0.39, 0.88]&+0.22 [$-$0.64, 0.98]&21/23&1,925/1,493\\
\mbox{LLMDet-L}&Q8&Box check&$-$0.32 [$-$0.91, 0.30]&$-$0.38 [$-$1.06, 0.33]&20/20&1,440/1,341\\
\mbox{LLMDet-L}&Q8&32-proposal rule&$-$0.21 [$-$0.64, 0.35]&$-$0.14 [$-$0.74, 0.53]&12/18&1,356/919\\
\midrule
\mbox{LLMDet-L}&Q16&raw names&+1.31 [0.70, 1.89]&+1.87 [1.12, 2.47]&53/28&1,785/1,559\\
\mbox{LLMDet-L}&Q16&NoM&+0.35 [$-$0.16, 0.73]&+0.62 [$-$0.01, 1.08]&45/31&1,758/1,364\\
\mbox{LLMDet-L}&Q16&Box check&$-$0.61 [$-$1.13, 0.09]&$-$0.55 [$-$1.04, 0.23]&47/29&1,296/1,322\\
\mbox{LLMDet-L}&Q16&32-proposal rule&$-$0.49 [$-$0.97, 0.15]&$-$0.27 [$-$0.75, 0.40]&24/24&1,228/901\\
\midrule
\mbox{LLMDet-L}&full80&raw names&+0.71 [0.47, 0.90]&+0.83 [0.55, 1.05]&285/263&--\\
\mbox{LLMDet-L}&full80&NoM&$-$0.08 [$-$0.27, 0.08]&+0.01 [$-$0.22, 0.19]&206/248&--\\
\mbox{LLMDet-L}&full80&Box check&$-$0.42 [$-$0.59, $-$0.22]&$-$0.18 [$-$0.42, 0.02]&193/186&--\\
\mbox{LLMDet-L}&full80&32-proposal rule&$-$0.15 [$-$0.31, 0.02]&$-$0.03 [$-$0.22, 0.15]&147/155&--\\
\bottomrule\end{tabular}
\end{table*}

\section{Learned name selectors: Presence and Utility}
\label{app:n2}
\supports{sec:limits}Learned name selectors do not beat the fixed concept rule. Presence learns whether a name is present, and Utility learns a name's measured effect under the query. Neither has a confirmed gain over NoM in any cell, and Utility needs three conditioned decodes instead of one. This is expected, since NoM is already within 0.69 and 2.44 AP of annotated present names at Q8 (Overview).
\begin{table*}[tbp]
\centering\small\setlength{\tabcolsep}{3pt}
\caption{\textbf{Utility absolute results on the 5,000 COCO validation images.} AP/AP50 and takeover rates have marginal image-bootstrap 95\% intervals. Matched/total concerns queried GT. TO counts taken-over unqueried GT. At full80 there are no unqueried GT, so the rate is not applicable (--).}
\label{tab:n2absolute}
\begin{tabular}{llllrrl}\toprule
Model&Query&AP [95\% CI]&AP50 [95\% CI]&Matched/total&TO&TO/100 [95\% CI]\\\midrule
\mbox{GD-L}&Q8&51.85 [50.82, 55.16]&68.12 [66.46, 71.43]&3,461/3,871&3,307&10.19 [9.59, 10.83]\\
\mbox{GD-L}&Q16&50.69 [50.21, 53.83]&66.54 [65.74, 69.75]&6,845/7,623&4,717&16.43 [15.63, 17.26]\\
\mbox{GD-L}&full80&50.32 [49.81, 51.61]&66.23 [65.52, 67.55]&32,104/36,335&0&--\\
\mbox{LLMDet-L}&Q8&57.05 [56.12, 60.31]&72.27 [70.89, 75.41]&3,713/3,871&7,789&23.99 [23.11, 24.88]\\
\mbox{LLMDet-L}&Q16&55.85 [55.35, 59.09]&71.38 [70.53, 74.56]&7,328/7,623&9,460&32.95 [31.95, 34.03]\\
\mbox{LLMDet-L}&full80&58.10 [57.55, 59.36]&75.10 [74.32, 76.28]&34,343/36,335&0&--\\
\bottomrule\end{tabular}
\end{table*}

\paragraph{Presence: presence confirmation.}
The frozen CLIP scorer ranks the 1,203 LVIS names and nominates 32. A frozen detector forward using these names without $Q$ supplies each name's maximum box score, number of boxes above $0.1$, top-three mean score, and maximum-score box area fraction. With CLIP score, rank, and LVIS training frequency, these seven features enter a $7\!\to\!64\!\to\!1$ MLP (577 trainable parameters). Separate confirmers are fitted for the two detectors. Training uses 20,000 hash-selected LVIS v1 training images from COCO train2017, with an 18,000/2,000 image split. Federated positive labels are 1, verified negatives 0, and unknown labels are excluded from binary cross-entropy. Adam uses learning rate $10^{-3}$ for 20 epochs. The held-out AUROC selects the epoch. Detectors and CLIP remain frozen.

After the concept filter, the scores rank names for conditioning only, and $k=4$ is fixed on the 2,000-image selection set without a probability threshold. The 32-proposal rule, a matched-input reference, uses threshold $0.2$ with the same $k=4$. Presence passes its \mbox{LLMDet-L} Q8 test over raw names (Sec.~\ref{app:mn}).

\paragraph{Utility: supervise what a name does under the query.}
A present name can still lower a queried score, as the direction rule allows (Eq.~\eqref{eq:rule}). Utility changes the target from presence to the measured effect of inserting a single name under $Q$. The label set is 1,500 hash-fixed LVIS training images, disjoint from COCO validation, with one image-hash Q8 per image. Candidates are the top 16 of the 32 proposed names, screened by the concept filter without backfilling. Each label needs one forward with $Q$ and one per candidate with $Q$ plus that name, with outputs still restricted to $Q$.

Let $S_{\rm TP}$ sum the highest correct queried-class score over queried GT, $C_{\rm TO}$ count taken-over unqueried objects, and $S_{\rm empty}$ sum the ten highest queried-report scores among boxes with zero intersection with every annotated COCO and LVIS object box, under the label-builder's fixed matching rules. $S_{\rm empty}$ is a conservative annotation-based proxy for background. The target, with both coefficients fixed before labeling, is
\begin{equation}
u(x,Q,n)=\Delta S_{\rm TP}-0.1\Delta C_{\rm TO}-0.1\Delta S_{\rm empty}.
\label{eq:n2utility}
\end{equation}
Ground truth enters only the training labels and is unavailable at test time. Like any surrogate, $u$ need not rank names as AP would.

Utility adds query-conditioned features to Presence's seven inputs: maximum and mean candidate-to-query text cosine, minimum candidate-to-query WordNet path distance, and maximum IoU between the candidate's highest-score box from the query-independent 32-proposal evidence and each queried class's highest-score box from a $Q$-only baseline. A two-layer MLP with hidden width 64 regresses $u$. A 10\% image-grouped holdout selects the epoch by Spearman correlation. After concept filtering, names with positive predicted utility are selected greedily, up to $k\in\{4,8\}$ fixed on the 2,000-image selection set. The design asks whether single-name utility suffices once query relatives are removed, so interactions between names are not modeled.

At test time Utility needs three conditioned decodes: the 32-proposal decode without $Q$, the $Q$ baseline for query-box features, and $Q$ plus the selected names. The image backbone is shared and the 32-proposal evidence can be cached, but the reported latency charges all three decodes and nomination. The output budget stays at 900 candidates and at most 300 queried reports.

Utility is compared with raw names, NoM, Box check, Presence, and the 32-proposal rule in both models at Q8, Q16, and full80. Only the \mbox{LLMDet-L} full80 test passes, and the joint requirement is not met.

\subsection{Query-conditioned single-name utility}
All six Utility cells (two models, three query sizes) use the 5,000 COCO validation images, the image-specific queries, the single-label top-300 endpoint, and 5,000 paired image resamples (seed 20261110). Training uses Q8 labels, so Q16 and full80 test transfer. Tables~\ref{tab:n2absolute} and~\ref{tab:n2contrasts} give absolute results, object accounting, and all paired references. The intervals reflect image variation and not training-seed variation.

\begin{table*}[tbp]
\centering\small\setlength{\tabcolsep}{3pt}
\caption{\textbf{Utility minus five references in every cell (pp).} 5,000 paired image resamples, seed 20261110. R/L counts recovered/lost queried GT. A/N counts avoided/new takeovers. The 32-proposal rule is the matched-input presence reference of Table~\ref{tab:ncontrasts}: it does not receive Utility's current-query baseline or 11-dimensional features and is not an equal-compute ablation. Values are rounded to two decimals. A/N is not applicable at full80 (--). No multiple-comparison correction is applied. Intervals do not include training-seed variation.}
\label{tab:n2contrasts}
\begin{tabular}{lllllrr}\toprule
Model&Query&Reference&$\Delta$AP [95\% CI]&$\Delta$AP50 [95\% CI]&R/L&A/N\\\midrule
\mbox{GD-L}&Q8&raw names&+0.18 [$-$0.59, 1.26]&$-$0.11 [$-$0.98, 1.08]&31/52&788/707\\
\mbox{GD-L}&Q8&NoM&$-$0.33 [$-$1.00, 0.58]&$-$0.47 [$-$1.25, 0.57]&27/51&795/663\\
\mbox{GD-L}&Q8&Box check&+0.69 [$-$0.24, 1.75]&+2.05 [0.83, 3.33]&56/108&2,806/627\\
\mbox{GD-L}&Q8&Presence&$-$0.28 [$-$1.00, 0.63]&+0.22 [$-$0.70, 1.30]&33/90&1,548/667\\
\mbox{GD-L}&Q8&32-proposal rule&+0.54 [$-$0.46, 1.48]&+1.35 [0.09, 2.41]&37/91&1,977/627\\
\midrule
\mbox{GD-L}&Q16&raw names&+0.35 [$-$0.41, 0.95]&+0.35 [$-$0.52, 1.05]&75/108&940/884\\
\mbox{GD-L}&Q16&NoM&$-$0.46 [$-$1.14, 0.05]&$-$0.40 [$-$1.16, 0.27]&66/112&984/764\\
\mbox{GD-L}&Q16&Box check&+0.99 [0.17, 1.58]&+2.25 [1.17, 2.93]&113/215&3,171/796\\
\mbox{GD-L}&Q16&Presence&+0.01 [$-$0.63, 0.48]&+0.55 [$-$0.27, 1.15]&85/154&1,757/843\\
\mbox{GD-L}&Q16&32-proposal rule&+1.01 [0.26, 1.62]&+1.74 [0.78, 2.41]&91/158&2,226/826\\
\midrule
\mbox{GD-L}&full80&raw names&+0.09 [$-$0.19, 0.37]&\phantom{+}0.00 [$-$0.34, 0.37]&344/382&--\\
\mbox{GD-L}&full80&NoM&$-$0.41 [$-$0.67, $-$0.15]&$-$0.49 [$-$0.82, $-$0.15]&270/404&--\\
\mbox{GD-L}&full80&Box check&$-$1.21 [$-$1.54, $-$0.92]&$-$1.56 [$-$2.00, $-$1.15]&319/827&--\\
\mbox{GD-L}&full80&Presence&$-$0.38 [$-$0.64, $-$0.12]&$-$0.61 [$-$0.96, $-$0.28]&332/559&--\\
\mbox{GD-L}&full80&32-proposal rule&$-$0.68 [$-$0.92, $-$0.41]&$-$0.99 [$-$1.29, $-$0.64]&339/649&--\\
\midrule
\mbox{LLMDet-L}&Q8&raw names&+1.85 [0.91, 2.65]&+2.21 [1.09, 3.02]&16/15&1,426/1,311\\
\mbox{LLMDet-L}&Q8&NoM&+0.11 [$-$0.53, 0.72]&+0.12 [$-$0.62, 0.77]&9/16&1,427/1,156\\
\mbox{LLMDet-L}&Q8&Box check&$-$0.48 [$-$1.21, 0.25]&$-$0.48 [$-$1.31, 0.40]&19/24&1,673/1,735\\
\mbox{LLMDet-L}&Q8&Presence&$-$0.16 [$-$0.93, 0.63]&$-$0.10 [$-$0.97, 0.81]&17/22&1,575/1,736\\
\mbox{LLMDet-L}&Q8&32-proposal rule&$-$0.38 [$-$1.07, 0.47]&$-$0.25 [$-$1.09, 0.69]&16/27&2,009/1,733\\
\midrule
\mbox{LLMDet-L}&Q16&raw names&+1.57 [0.80, 2.20]&+1.99 [1.06, 2.65]&39/24&1,282/1,309\\
\mbox{LLMDet-L}&Q16&NoM&+0.61 [$-$0.13, 1.12]&+0.75 [$-$0.09, 1.29]&31/27&1,256/1,115\\
\mbox{LLMDet-L}&Q16&Box check&$-$0.35 [$-$1.05, 0.45]&$-$0.42 [$-$1.15, 0.48]&53/45&1,488/1,767\\
\mbox{LLMDet-L}&Q16&Presence&+0.26 [$-$0.45, 0.87]&+0.13 [$-$0.67, 0.80]&31/41&1,433/1,686\\
\mbox{LLMDet-L}&Q16&32-proposal rule&$-$0.23 [$-$0.93, 0.59]&$-$0.14 [$-$0.93, 0.73]&35/45&1,790/1,716\\
\midrule
\mbox{LLMDet-L}&full80&raw names&+0.89 [0.65, 1.09]&+0.82 [0.56, 1.06]&235/185&--\\
\mbox{LLMDet-L}&full80&NoM&+0.10 [$-$0.09, 0.28]&\phantom{+}0.00 [$-$0.20, 0.21]&166/180&--\\
\mbox{LLMDet-L}&full80&Box check&$-$0.24 [$-$0.45, 0.01]&$-$0.18 [$-$0.44, 0.07]&254/219&--\\
\mbox{LLMDet-L}&full80&Presence&+0.18 [$-$0.01, 0.39]&\phantom{+}0.00 [$-$0.22, 0.25]&234/206&--\\
\mbox{LLMDet-L}&full80&32-proposal rule&+0.04 [$-$0.16, 0.26]&$-$0.03 [$-$0.27, 0.22]&237/217&--\\
\bottomrule\end{tabular}
\end{table*}

The \mbox{LLMDet-L} Q8 test over Presence is not met, with an AP50 change of $-0.102$ $[-0.974,0.811]$. Against NoM, Box check, and the 32-proposal rule, the AP50 changes are $+0.123$ $[-0.618,0.771]$, $-0.482$ $[-1.314,0.395]$, and $-0.245$ $[-1.093,0.689]$, and the AP intervals also cross zero. At full80, Utility changes \mbox{GD-L} AP50 over raw names by $0.000$ $[-0.338,0.374]$ and misses its test, while \mbox{LLMDet-L} gains $0.823$ $[0.561,1.062]$ and passes. One test of three passing leaves the joint requirement unmet and does not show that native full80 accuracy is preserved.

Utility does improve on the weaker references. On \mbox{LLMDet-L} it exceeds raw names in AP and AP50 at all three query sizes, and on \mbox{GD-L} its Q8/Q16 AP50 and Q16 AP exceed Box check and the 32-proposal rule. No AP or AP50 gain over NoM or Presence is confirmed in any cell, and \mbox{GD-L} full80 is lower than NoM, Box check, Presence, and the 32-proposal rule. For \mbox{LLMDet-L} Q8 against Presence, Utility recovers 17 and loses 22 queried GT, and it avoids 1,575 takeovers while adding 1,736.

Utility and Presence differ in target, features, training set, selection, candidate handling, and computation, and Utility adds a current-query baseline, 11-dimensional features, and a third decode that the 32-proposal rule lacks. The comparison is between whole methods, so it neither isolates utility supervision nor equalizes compute.

On held-out labels, predicted and measured utility reach Spearman $0.283$ (\mbox{GD-L}, epoch 19) and $0.291$ (\mbox{LLMDet-L}, epoch 18), with 1,350 fit and 150 holdout images. These correlations only select the epoch. The charged component sums are 1,339.99 and 1,377.76 ms/image at Q8 for \mbox{GD-L} and \mbox{LLMDet-L}, and 1,627.16 and 1,667.40 ms/image at full80. They include the 32-proposal and current-query decodes, CPU features and MLP, the selected conditioning, and the CLIP and image cost, and they exclude setup, so they are component sums and not end-to-end wall-clock times. When no extra name is selected, the baseline decode is reused and not charged twice.

\section{Score decomposition of the withdrawal effect}
\label{app:score_accounting}
\supports{sec:measurement}These definitions keep the conditioning effect of Eq.~\eqref{eq:fixed} separate from output competition and from changes in the class text. The controlled withdrawal study of Sec.~\ref{sec:measurement} uses 400 development images and 587 targets (577 for GLIP), the top 20 predictions, and no score floor.
For $\Delta_{\rm fixed}$, the own-minus-absent takeover differences in percentage points on \mbox{GD-L} are $15.84$ $[12.27,19.40]$ at decoder text attention, $1.19$ $[-0.69,3.21]$ at feature enhancement, and $0.51$ $[-0.55,1.67]$ at query selection. In GLIP the late three fusion blocks give $22.49$ $[18.57,26.40]$, and the first three give $0.68$ $[-0.89,2.26]$. The fixed-box \mbox{YOLO-World} neck gives $12.78$ $[10.00,15.63]$. OWLv2 gives zero.

\subsection{Matched total and fixed-readout contrasts}
For a fixed-readout contrast $\Delta_{\rm fixed}$ (Eq.~\eqref{eq:fixed}), the corresponding native end-to-end contrast is $\Delta_{\rm total}$. Only when image IDs, physical targets, output definition, and threshold all coincide do we form
\begin{equation}
\Delta_{\rm residual}=\Delta_{\rm total}-\Delta_{\rm fixed}.
\label{eq:residual}
\end{equation}
This operational residual includes output-competition effects and any interaction with the changed readout. It is not automatically an identified natural direct effect. Ratios $\Delta_{\rm fixed}/\Delta_{\rm total}$ are omitted when the denominator is zero or unstable. Confidence intervals must resample the two components jointly by image. Intervals from unrelated experiments cannot be subtracted to manufacture a paired interval.

\subsection{Visual, textual, and cross contributions}
The pre-specified score decomposition compares the same physical candidate and the same class-text coordinates. If a raw bilinear score is $s=h^\top t$, with state changes $\delta h$ and $\delta t$, then
\begin{equation}
\delta s=(\delta h)^\top t+h^\top\delta t+(\delta h)^\top\delta t.
\label{eq:bilinear}
\end{equation}
The three terms are visual-side, text-side, and cross contributions. Their sum must match the measured raw score difference to numerical precision before any proportion is reported. Normalization, temperature, bias, sigmoid, and changing candidate indices require explicit treatment. Equation~\eqref{eq:bilinear} does not silently absorb them. A visual proportion is undefined when its selected denominator vanishes and need not lie between zero and one when terms cancel. A changed box covering the same object is not sufficient evidence that candidate features share the same coordinate system.

\section{Visual comparisons and exact reference tables}
\label{app:visual_comparisons}
\supports{sec:main}This section gives the definitions, controls, and exact values behind the main-paper figures.
For $N$ eligible absent-query images, let $F_i$ indicate that image $i$ has at least one reported false positive. The commission cost used throughout is
\begin{equation}
 C_{\rm commission}=\frac{100}{N}\sum_{i=1}^{N}F_i.
\label{eq:commission}
\end{equation}

\subsection{Paired inference-control comparisons}
\label{app:control_details}
Raw names exceed sinks, equal-count random names, and CLIP veto on both detectors at Q8. Without per-detector tuning, they come within 0.72 AP50 of a development-tuned fixed-name reference on \mbox{GD-L} and exceed it on \mbox{LLMDet-L}.
The raw-name AP50 changes on GD-Tiny and GLIP are $-0.11$ $[-1.93,1.39]$ and $+3.83$ $[2.13,5.68]$, respectively. The GLIP gain is not confirmed on the 833 COCO images outside the M-OWODB test set. These raw-name transfer endpoints are distinct from concept-filtered NoM.

\paragraph{Sinks and stronger inference controls.}
The sink-only reference improves \mbox{GD-L} AP50 over native by $7.98$ $[6.41,9.40]$ at Q8 and $5.86$ $[4.62,6.89]$ at Q16. \mbox{LLMDet-L} gains $3.10$ $[1.88,3.93]$ and $1.80$ $[1.25,2.27]$. The raw-name baseline exceeds this sink-only reference in both models and both query sizes. At Q8 the paired margins are $4.46$ $[2.15,5.49]$ and $3.39$ $[1.84,4.72]$. Relative to equal-count random names, the raw-name baseline's Q8 margins are $4.36$ $[2.84,5.65]$ and $3.09$ $[1.37,4.12]$. Against CLIP veto, they are $16.45$ $[13.03,17.79]$ and $9.48$ $[8.01,11.35]$.

Fixed context is a strong practical alternative on \mbox{GD-L}, whereas image-proposed context contributes more on \mbox{LLMDet-L}. The raw-name baseline minus the fixed-name-plus-sink reference is $-0.72$ $[-2.08,0.33]$ at Q8 and $-0.77$ $[-1.61,0.12]$ at Q16 for \mbox{GD-L}, versus $3.29$ $[1.16,4.23]$ and $1.00$ $[-0.13,2.18]$ for \mbox{LLMDet-L}. That reference combines fixed names and sinks, so this tests a practical alternative and does not isolate image adaptivity alone.

\subsection{Open-world transfer details}
\label{app:ow_details}

On \mbox{GD-L}, NoM improves known AP50 and lowers open-set error at both stages (Table~\ref{tab:owlegal}). A separate 80-known-class diagnostic loses $1.31$ AP50 points in \mbox{GD-L} with an open proposal bank. It is not part of the legal-bank comparison below.

The M-OWODB test images were also used during development. We use the frozen \mbox{GD-L}, the authors' 100-report budget, known scores above $0.05$, and a legal 874-name bank with all pre-specified future-class exclusions. All four arms output only known classes. Absolute open-set error (A-OSE) counts known-class reports matched to unknown objects. Multiple reports on one unknown object can contribute multiple errors.
\begin{table}[tbp]
\centering\small\setlength{\tabcolsep}{2.8pt}
\caption{\textbf{Known-only M-OWODB test set.} Each cell is known AP50 / A-OSE. The object token, proposed names, and sinks are conditioning only. Unknown AP and recall are structurally zero in these arms.}
\label{tab:owlegal}
\begin{tabular}{lrr}\toprule
Condition & 20 known & 40 known\\\midrule
Native &72.77 / 27,589&61.74 / 17,805\\
Object &75.40 / 18,988&63.95 / 10,548\\
NoM &73.74 / 17,543&63.04 / 10,179\\
NoM + 24 sinks &74.08 / 13,036&63.07 / 7,569\\\bottomrule
\end{tabular}
\end{table}
The combination minus NoM gains $0.34$ $[0.17,0.51]$ AP50 at 20 classes and $0.03$ $[-0.44,0.33]$ at 40. A-OSE drops by 4,507 and 2,610, respectively, with both paired intervals excluding zero. Against NoM, the combination recovers 81 and loses 206 known objects at 20 classes and recovers 157 and loses 327 at 40, so its AP gain is not an improvement in recall. The object reference has higher known AP50 at both stages. Intervals use 1,000 paired integer-ID-cluster draws with a fixed tie order. They are not interchangeable with the 5,000-draw COCO intervals.
The same conditioning operation transfers differently to \mbox{LLMDet-L}. NoM reduces A-OSE at both stages but changes known AP50 by $-3.63$ $[-4.27,-3.01]$ at 20 classes and $-1.87$ $[-2.31,-1.37]$ at 40. The combination recovers $0.95$ $[0.76,1.18]$ and $0.55$ $[0.24,0.82]$ points over NoM, yet remains below native AP50. The stage-20 joint test is not met. These results separate reduced misreporting from improved recognition. All arms and both stages appear in Sec.~\ref{app:a3_extension}.

\subsection{Expanded protocol, query-draw, and ranking details}
\label{app:expanded_details}
The matched controls include equal-count random names, a frozen image-independent vocabulary plus 24 sinks, and CLIP veto at the threshold $0.5$ fixed on the selection set. The sink-only reference appends 24 BERT unused tokens initialized with the frozen word-embedding table mean, with no extra context names. These tokens traverse the native language encoder, fusion, query selection, and decoder. They are excluded before output argmax. The fixed-context reference uses the same sinks with four frozen names. Thus the sink-only reference is a whole-path conditioning baseline, not an intervention restricted to the last decoder attention layer. This implementation is evaluated under our common reporting interface, not the Tractor-Dogs benchmark itself.

NoM fits no parameters, so we vary the queried class sets and not training seeds. Under the main draw and two additional fixed hash salts, Q8 gains over native are 12.80, 13.02, and 13.14 AP50 for \mbox{GD-L}, and 8.58, 8.13, and 8.42 for \mbox{LLMDet-L}. Their means and sample standard deviations are $12.99\pm0.17$ and $8.37\pm0.23$. Omission costs are $16.60\pm0.34$ and $13.45\pm0.81$. All three draws are retained. These standard deviations measure query-sampling variation, whereas the paired intervals measure image variation. Neither estimates the training-seed variance of learned controls. Section~\ref{app:robustness} adds two further draws.

For the annotation-assisted comparison in Table~\ref{tab:oracle}, paired AP50 gains from adding present names over the short prompt are $12.43$ $[10.28,13.52]$ on \mbox{GD-L} and $11.04$ $[9.21,12.58]$ on \mbox{LLMDet-L}. Present minus absent gives $11.78$ $[9.88,12.93]$ and $9.85$ $[8.40,11.06]$, respectively. Full conditioning minus present-only gives $4.13$ $[2.84,5.10]$ and $2.10$ $[0.80,2.91]$.

Table~\ref{tab:rankscale} compares the 11 checkpoints, including HDINO~\cite{hdino2026}, OV-DEIM~\cite{ovdeim2026}, and YOLOE~\cite{yoloe2025}. The full80--Q8 rank correlation is $0.855$, also $0.855$ when the full-prompt reference is evaluated on the same Q8 classes. Besides LLMDet--WeDetect, GD-B--WeDetect reverses in point ordering. \mbox{GD-B} minus WeDetect is $+2.55$ AP50 at full80 and $-0.61$ $[-2.44,0.90]$ at Q8. Its same-Q8 full-prompt margin is $+2.18$ $[0.95,3.89]$. The extra reversal is descriptive because its Q8 interval includes zero.

\begin{figure}[tbp]
\centering\includegraphics[width=\linewidth]{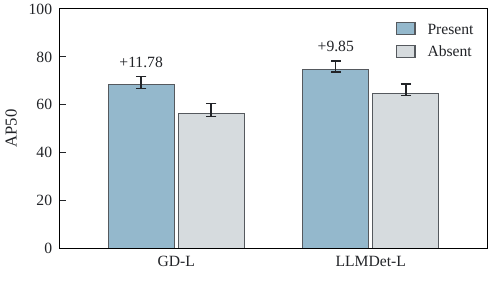}
\caption{\textbf{Present names recover accuracy without becoming answers.} Q8, 5,000 COCO validation images, Q-only output. Bars show AP50 with 95\% image-bootstrap intervals. Numbers above the present-name bars give present-minus-absent gains. Their paired intervals are reported in the text. Names use annotations and are not a deployed method.}
\label{fig:present}
\end{figure}
\begin{figure}[tbp]
\centering\includegraphics[width=\linewidth]{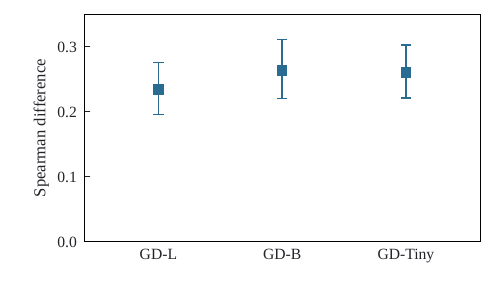}
\caption{\textbf{Signed response predicts withdrawal better than mass.} Points show the Spearman correlation of signed first-order response minus that of own-name attention mass, with paired 95\% intervals. The comparison concerns controlled full-withdrawal score changes, not NoM gains. The GD records are the same matched set visualized in Sec.~\ref{app:added_evidence}. The GLIP layer-wise records use a separate protocol.}
\label{fig:direction}
\end{figure}
\begin{figure*}[tbp]
\centering\includegraphics[width=\textwidth]{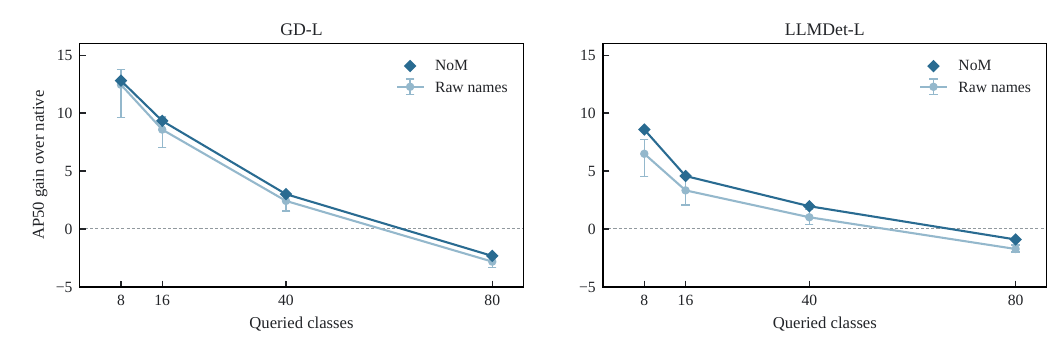}
\caption{\textbf{Context gains fall as the query grows.} The Q40 follow-up reused the 5,000 COCO val2017 images, per-image queries, frozen NoM settings, and Q-only readout. AP50 gains over native were 3.00 [2.24, 3.63] for \mbox{GD-L} and 1.96 [1.35, 2.62] for \mbox{LLMDet-L}. Gains over Raw names were 0.59 [0.31, 0.97] and 0.95 [0.70, 1.20], respectively. All 95\% CIs use 5,000 paired image resamples.}
\label{fig:curve}
\end{figure*}
\begin{figure}[tbp]
\centering\includegraphics[width=\linewidth]{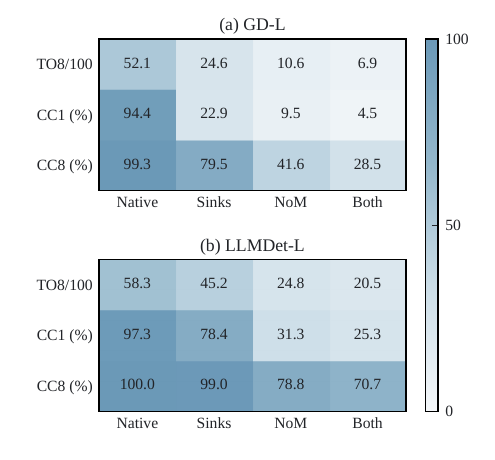}
\caption{\textbf{Names and sinks jointly reduce the recorded errors.} \mbox{GD-L} (a) and \mbox{LLMDet-L} (b). All values are lower-is-better, on a shared 0--100 color scale (darker means more errors). TO8/100 counts taken-over GT per 100 unqueried GT. CC1/CC8 count percentages of images with any false report under absent queries. Units are not pooled. All 5,000 images are retained. TO8 has 32,464 eligible GT. Cells are rounded to one decimal. Exact reported values are in Table~\ref{tab:commission}. Lower error counts do not guarantee higher AP.}
\label{fig:error_matrix}
\end{figure}

\begin{table*}[tbp]
\centering\small\setlength{\tabcolsep}{3pt}
\caption{\textbf{Partial queries change the measured ordering of checkpoints.} COCO val2017, 5,000 images. Parentheses are point-estimate ranks in the full80 and Q8 columns. Full80 evaluates all classes. Full$\to$Q8 and Q8 evaluate the same image-level query classes. Omission cost is full$\to$Q8 minus Q8, with 5,000 paired image resamples. This is our single-label reporting interface, not a reproduction of every model's leaderboard. The NoM column uses the unchanged NoM recipe and does not enter native rankings. Dashes denote unmeasured cells. Candidate counts, NMS, text language, training data, and checkpoints differ (Sec.~\ref{app:a1_details}).}
\label{tab:rankscale}
\begin{tabular}{lrrrrr}\toprule
Checkpoint & Full80 AP50 (rank) & Full$\to$Q8 AP50 & Q8 AP50 (rank) & Omission cost at Q8 [95\% CI] & NoM Q8 AP50\\\midrule
LLMDet & 76.01 (1) & 76.50 & 63.57 (4) & +12.93 [10.71, 14.22] & 72.15\\
\mbox{GD-B} & 73.46 (2) & 75.41 & 72.44 (2) & +2.96 [1.94, 4.02] & 67.64\\
WeDetect & 70.91 (3) & 73.22 & 73.05 (1) & +0.17 [$-$0.79, 0.56] & --\\
\mbox{GD-L} & 69.06 (4) & 71.99 & 55.78 (7) & +16.21 [13.59, 17.35] & 68.58\\
GD-Tiny & 65.79 (5) & 68.43 & 59.49 (5) & +8.94 [7.32, 10.40] & 61.03\\
\mbox{HDINO-T} & 62.30 (6) & 65.62 & 66.78 (3) & $-$1.16 [$-$1.78, 0.83] & 64.67\\
\mbox{OV-DEIM-S} & 55.25 (7) & 58.01 & 58.74 (6) & $-$0.73 [$-$1.71, $-$0.11] & 57.12\\
OWLv2 & 50.44 (8) & 50.45 & 50.82 (8) & $-$0.37 [$-$1.37, 0.42] & --\\
YOLOE & 45.77 (9) & 47.59 & 47.76 (9) & $-$0.17 [$-$0.75, 0.41] & --\\
\mbox{GLIP-TB} & 28.85 (10) & 33.90 & 30.04 (10) & +3.86 [2.40, 6.24] & 34.08\\
YOLO-World & 11.84 (11) & 19.40 & 18.43 (11) & +0.97 [0.12, 1.74] & 19.24\\
\bottomrule\end{tabular}
\end{table*}

\begin{table}[tbp]
\centering\small\setlength{\tabcolsep}{4pt}
\caption{\textbf{Annotation-assisted context comparison at eight queried classes.} Official AP/AP50, same 5,000 images and Q-only output. Additional context is ineligible as output. Full conditioning is an upper reference under this interface, not a universally attainable deployment input.}
\label{tab:oracle}
\begin{tabular}{lrrrr}\toprule
 &\multicolumn{2}{c}{GD-L}&\multicolumn{2}{c}{LLMDet-L}\\
Condition & AP&AP50&AP&AP50\\\midrule
Short &44.26&55.78&51.21&63.57\\
+ present names &52.87&68.21&59.38&74.62\\
+ absent names &44.77&56.43&52.22&64.76\\
Full condition, Q-only &55.87&72.34&60.94&76.71\\\bottomrule
\end{tabular}
\end{table}

\begin{table}[tbp]
\centering\small\setlength{\tabcolsep}{2.8pt}
\caption{\textbf{Present-object takeovers and commission cost.} Native / 24 sinks / NoM / both, lower is better. TO8/100 is taken-over GT per 100 unqueried GT at Q8. CC1/CC8 are percentages of images with a false report when querying 1/8 absent classes. Each model uses all 5,000 images. TO8 has 32,464 eligible GT.}
\label{tab:commission}
\begin{tabular}{llrrrr}\toprule
Model & Error & Native & Sinks & NoM & Both\\\midrule
\mbox{GD-L} & TO8/100 &52.09&24.57&10.59&6.92\\
\mbox{LLMDet-L} & TO8/100 &58.30&45.22&24.83&20.53\\
\mbox{GD-L} & CC1 (\%) &94.42&22.92&9.52&4.48\\
\mbox{LLMDet-L} & CC1 (\%) &97.30&78.44&31.32&25.34\\
\mbox{GD-L} & CC8 (\%) &99.30&79.52&41.60&28.48\\
\mbox{LLMDet-L} & CC8 (\%) &100.00&99.04&78.80&70.72\\\bottomrule
\end{tabular}
\end{table}

\paragraph{Qualitative filter pairs.}
The train pair in Fig.~\ref{fig:class_differences} and the additional \mbox{LLMDet-L} cases below were selected from stored predictions after evaluation. COCO image 70254 (Q8) removes \emph{passenger car part of a train}, \emph{railcar part of a train}, \emph{train railroad vehicle}, and \emph{bullet train}, retaining \emph{cabin car}, \emph{freight car}, \emph{lime}, and \emph{bowler hat}. The displayed train reports have scores 0.148 and 0.898. Image 285788 (Q16) removes \emph{ostrich} and retains \emph{giraffe, zebra, rhinoceros, elephant, gazelle, playpen, cow}. The displayed bird report appears with score 0.390. Image 31248 (full80) removes \emph{armchair, recliner, chaise longue, rocking chair} and retains \emph{bookcase, fireplace, footstool, armoire}. The displayed chair report appears with score 0.512. Queries are identical within each pair. Filtered lists are subsets with no refill. Both predictions use the pre-specified single-label, 300-report, score $>0.05$ interface. Ground truth is used only for selecting examples after evaluation and for matching at IoU $\geq0.5$, never for proposal generation or filtering. Before/after crops use identical coordinates. All drawn boxes come from stored predictions. Changes to multiple names also change text length, so these examples do not separate lexical identity from length or interactions.

\section{Additional comparisons for the main claims}
\label{app:added_evidence}

\supports{sec:main}Four paired comparisons complete the main claims. Sinks leave 94\% of the own-name effect in place, the omission cost predicts the gain of NoM, \mbox{LLMDet-L} with NoM exceeds WeDetect in AP, and on \mbox{GD-L} NoM comes within 0.02 AP of a development-tuned fixed-name reference.

\paragraph{Attention sinks and the own-name effect.}
We repeated the fixed-readout own-name measurement of Sec.~\ref{sec:measurement} with 24 mean-initialized sink rows appended to decoder text attention only (\mbox{GD-L}, the same 400 development images and 587 targets, top 20 predictions, no score floor). The own-minus-absent effect is $15.84$ $[12.27,19.40]$ points without sinks and $14.82$ $[11.37,18.24]$ with them, a paired change of $-1.02$ $[-2.20,0.004]$. With a $0.25$ score floor the change is $-2.21$ $[-3.59,-1.00]$, and with the top 100 predictions it is $0.00$ $[-0.50,0.50]$. In this decoder-only intervention, sink rows leave 94\% of the primary own-name effect in place. For detection AP, Sec.~\ref{sec:main} separately compares NoM with whole-path sinks under the common query interface.

\begin{table*}[t]
\centering\small\setlength{\tabcolsep}{5pt}
\caption{\textbf{Additional paired Q8 comparisons.} NoM minus each reference on 5,000 COCO images. Intervals use the same 5,000 paired image draws and the same queried classes. The first row compares different detector architectures. The lower panel retains the complete fixed-four-word-plus-24-sink reference. Notation follows Table~\ref{tab:main}.}
\label{tab:additional_paired_N}
\begin{tabular}{lrr}\toprule
Reference & $\Delta$AP [95\% CI] & $\Delta$AP50 [95\% CI]\\\midrule
\shortstack[l]{WeDetect\\(\mbox{LLMDet-L} NoM)} & $+1.48\,[0.04,3.17]$ & $-0.90\,[-2.61,0.97]$\\
\shortstack[l]{Fixed+sinks\\(\mbox{GD-L} NoM)} & $-0.02\,[-1.03,0.74]$ & $-0.37\,[-1.59,0.58]$\\\bottomrule
\end{tabular}
\par\smallskip
\begin{tabular}{lrrrrrr}\toprule
Model & Q8 AP ($\Delta$) & Q8 AP50 & Q16 AP50 & full80 AP & TO8/100 & $t_{\rm Q8}$ (ms)\\\midrule
\mbox{GD-L} &52.19\;(+7.93)&68.95&66.95&51.38&15.37&614.18\textsuperscript{$\Sigma$}\\
\mbox{LLMDet-L} &53.45\;(+2.24)&66.78&68.39&58.93&43.82&590.89\textsuperscript{$\Sigma$}\\
\bottomrule
\end{tabular}
\end{table*}

\begin{figure}[tbp]
\centering\includegraphics[width=\linewidth]{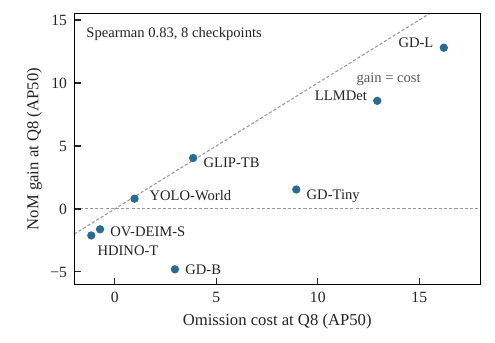}
\caption{\textbf{The gain of NoM follows the omission cost.} Each point is one text-conditioned checkpoint. The diagonal marks full recovery, and the horizontal dashed line marks zero.}
\label{fig:cost_gain}
\end{figure}

\paragraph{Omission cost predicts the gain of NoM.}
Figure~\ref{fig:cost_gain} plots the native Q8 omission cost against the Q8 AP50 gain of the unchanged NoM recipe for the eight text-conditioned checkpoints evaluated with NoM (values from Table~\ref{tab:rankscale}). The Spearman correlation is $0.83$, with an exact one-sided permutation $p=0.0077$ over all $8!$ orderings. Checkpoints without an omission cost (\mbox{HDINO-T}, \mbox{OV-DEIM-S}) lose a little, and \mbox{GD-B}, whose cost is small, loses most. GD-Tiny lies below the trend despite a large cost, consistent with the proposal-quality limit in Sec.~\ref{sec:limits}. With eight checkpoints, the relation serves as a rank indicator of when NoM is worth applying, not as a calibrated slope.

\paragraph{NoM and the strongest partial-query checkpoint.} Table~\ref{tab:additional_paired_N} summarizes this and the following paired comparison.
At Q8, \mbox{LLMDet-L} with NoM obtains $56.94$ AP versus WeDetect's $55.46$: the paired difference is $+1.48$ $[0.04,3.17]$. The AP50 difference is $-0.90$ $[-2.61,0.97]$, which establishes neither a difference nor equivalence. Comparisons share the same 5,000 images and per-image queries, reusing the same 5,000 bootstrap weights and excluding the unit-weight point from quantiles. Candidate counts and compute differ across detectors.

\paragraph{NoM and fixed names plus sinks on Grounding \mbox{DINO-L}.}
NoM minus fixed names plus sinks changes Q8 AP by $-0.02$ $[-1.03,0.74]$ and AP50 by $-0.37$ $[-1.59,0.58]$. Absolute AP/AP50 is $52.18/68.58$ versus $52.19/68.95$. The intervals establish neither a difference nor equivalence. These compare complete recipes, not image adaptivity alone.

\begin{figure*}[p]
\centering\includegraphics[width=\textwidth,trim=0bp 10bp 0bp 4bp,clip]{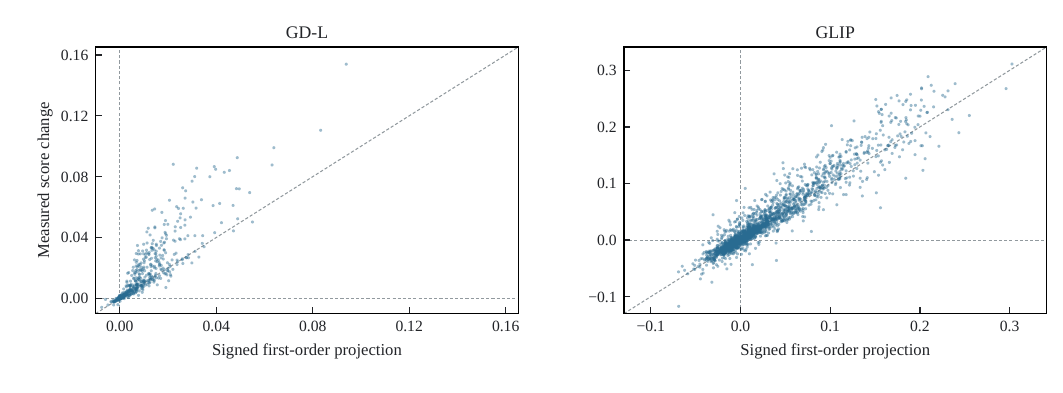}
\caption{\textbf{Earlier signed-response predictions.} All 587 \mbox{GD-L} records and 3,462 repeated GLIP layer responses. The latter are not independent objects. Diagonals mark equality, and horizontal and vertical dashed lines mark zero. \mbox{GD-L} uses the object set of Fig.~\ref{fig:direction}. GLIP uses a separate layer-wise protocol.}
\label{fig:historical_scatter}
\vspace{3pt}
\centering\includegraphics[width=\textwidth,trim=0bp 10bp 0bp 4bp,clip]{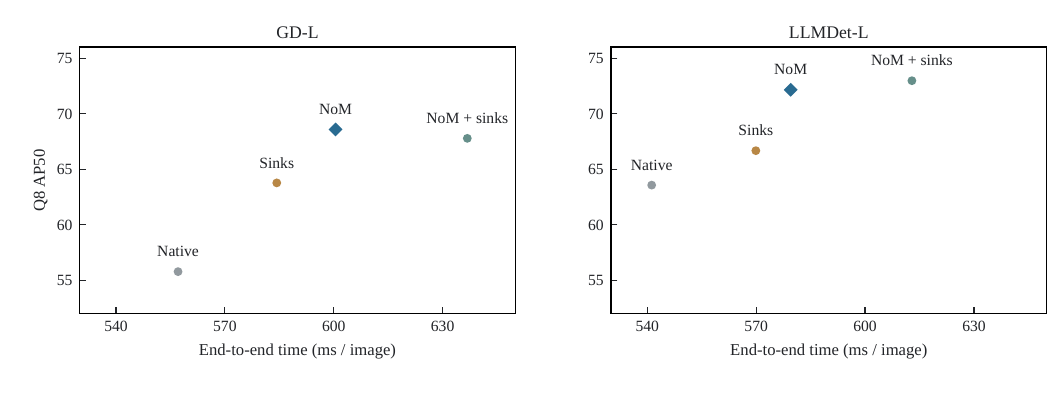}
\caption{\textbf{Measured end-to-end cost.} Four co-resident arms on the same GPU, timed on 100 images after 10 warm-ups, including nomination. Stage-sum rows (Table~\ref{tab:main}) are excluded from this axis.}
\label{fig:latency_plot}
\vspace{3pt}
\centering\includegraphics[width=\textwidth,trim=0bp 10bp 0bp 4bp,clip]{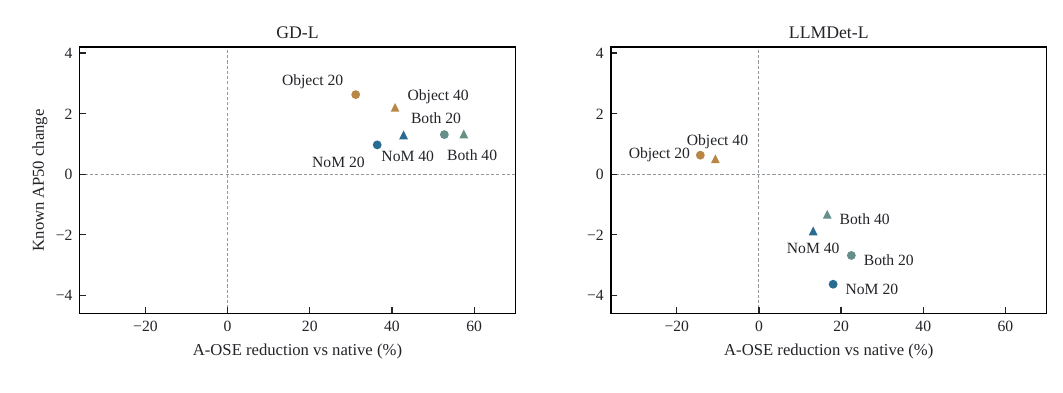}
\caption{\textbf{Known-only legal-bank transfer.} Object, NoM, and combined arms: relative A-OSE reduction versus known AP50 change. Circles/triangles denote 20/40 known classes. Models are frozen, and unknown recall is zero by design.}
\label{fig:owod_plot}
\end{figure*}

\begin{figure}[tbp]
\centering\includegraphics[width=\linewidth]{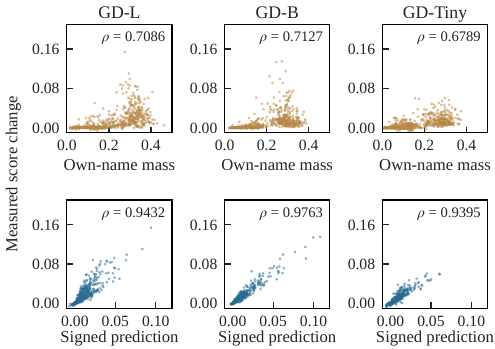}
\caption{\textbf{Direction predicts controlled score changes better than attention mass.} Three checkpoints, 587 matched objects each. Top: own-name attention share. Bottom: signed first-order response. Both rows use the identical measured own-minus-absent full-withdrawal score change of the fixed competing class. All points are shown, including negative values. Correlations are computed from the displayed records.}
\label{fig:matched_mechanism_N}
\end{figure}

\paragraph{Matched object-level mechanism comparison.}
Figure~\ref{fig:matched_mechanism_N} compares predictors on identical objects, queries, competing classes, and full-withdrawal endpoints: all 587 objects from 400 development images per checkpoint. Own-name mass gives Spearman $0.7086$, $0.7127$, and $0.6789$ for \mbox{GD-L}, \mbox{GD-B}, and GD-Tiny. Signed response gives $0.9432$, $0.9763$, and $0.9395$. The paired correlation differences are $0.2346$ $[0.1959,0.2759]$, $0.2636$ $[0.2201,0.3116]$, and $0.2606$ $[0.2215,0.3027]$, respectively. This tests controlled score changes, not NoM gains or natural-query takeover. Figure~\ref{fig:historical_scatter} shows the earlier \mbox{GD-L} and GLIP records.

\begin{figure}[tbp]
\centering\includegraphics[width=\linewidth]{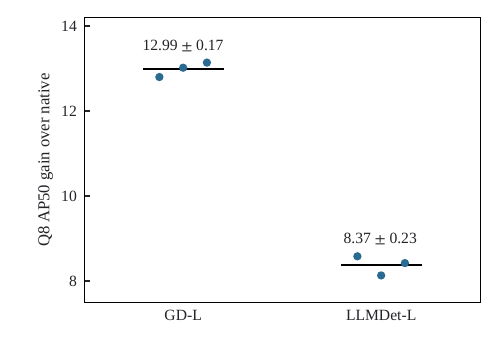}
\caption{\textbf{All three fixed query draws.} Each dot is one NoM-minus-native Q8 AP50 gain. Horizontal segments mark means. Labels report mean and sample SD from the eight-decimal measurement record. This measures query-set variation, not training-seed variation or a confidence interval.}
\label{fig:draw_plot}
\end{figure}

\begin{table*}[tbp]\centering\small\setlength{\tabcolsep}{3pt}
\caption{\textbf{NoM gains hold across query draws.} NoM minus native at Q8, COCO val2017, 5,000 images. Draws 1--3 are those of Sec.~\ref{sec:main}, rerun on the GPU of this section. Draws 4 and 5 are new.}
\label{tab:draws5}
\begin{tabular}{llrrr}\toprule
Model & Draw & $\Delta$AP [95\% CI] & $\Delta$AP50 [95\% CI] & R/L\\\midrule
\mbox{GD-L} & 1 & +7.91 [5.88, 9.13] & +12.79 [10.05, 14.02] & 91/140\\
 & 2 & +7.84 [5.72, 8.54] & +13.00 [9.96, 13.78] & 55/104\\
 & 3 & +8.30 [6.41, 9.32] & +13.14 [10.50, 14.28] & 62/119\\
 & 4 & +8.61 [6.48, 9.81] & +13.76 [10.99, 15.15] & 76/132\\
 & 5 & +7.53 [5.65, 8.70] & +12.22 [9.75, 13.58] & 56/121\\\midrule
\mbox{LLMDet-L} & 1 & +5.72 [4.16, 6.72] & +8.58 [6.65, 9.71] & 20/34\\
 & 2 & +5.54 [3.90, 6.75] & +8.13 [5.97, 9.47] & 22/24\\
 & 3 & +5.92 [4.35, 6.92] & +8.41 [6.31, 9.45] & 16/30\\
 & 4 & +6.02 [4.27, 7.00] & +9.00 [6.77, 10.00] & 20/32\\
 & 5 & +5.74 [4.20, 7.26] & +8.28 [6.25, 9.83] & 23/20\\\bottomrule
\end{tabular}
\end{table*}

\begin{table*}[tbp]\centering\small\setlength{\tabcolsep}{3pt}
\caption{\textbf{Nomination budget.} Paired differences against $k=8$ on the same 5,000 COCO images and Q8 queries. The intervals do not establish equivalence or show that $k=8$ is optimal.}
\label{tab:nom_k_budget}
\begin{tabular}{lrrrrrr}\toprule
Model & $k$ & AP & AP50 & $\Delta$AP [95\% CI] & $\Delta$AP50 [95\% CI] & R/L\\\midrule
\mbox{GD-L} & 4 & 52.17 & 67.90 & $-$0.01 [$-$0.66, 0.67] & $-$0.68 [$-$1.45, 0.22] & 48/21\\
 & 8 & 52.18 & 68.58 & -- & -- & --\\
 & 16 & 51.79 & 67.92 & $-$0.39 [$-$0.95, 0.17] & $-$0.66 [$-$1.42, 0.07] & 23/81\\\midrule
\mbox{LLMDet-L} & 4 & 57.27 & 72.27 & +0.33 [$-$0.29, 0.80] & +0.11 [$-$0.65, 0.70] & 18/16\\
 & 8 & 56.94 & 72.15 & -- & -- & --\\
 & 16 & 57.01 & 72.35 & +0.07 [$-$0.72, 0.67] & +0.20 [$-$0.72, 0.81] & 15/20\\\bottomrule
\end{tabular}
\end{table*}

\section{Robustness of the fixed recipe}
\label{app:robustness}
\supports{sec:main}The unchanged NoM recipe keeps its gain over native across five query draws, two CLIP proposers, and three nomination budgets on both detectors, and on a fixed eight-class LVIS minival query for \mbox{GD-L}. The query draws, the second proposer, the Q16 sink run, and the LVIS run were specified before their outcomes were seen, and every outcome is reported. The budget runs predate them and serve as a sensitivity analysis. No result here was used to change the recipe.

The COCO runs in this section use one GPU in strict FP32 and recompute their native baselines on that GPU. Their values differ from Table~\ref{tab:main} and Sec.~\ref{sec:main} by at most 0.02 points. Intervals use 5,000 paired image resamples. R/L counts queried GT recovered/lost at IoU 0.5 under the fixed report rule.

\paragraph{Query draws.}
Table~\ref{tab:draws5} adds two new hash salts to the three draws of Sec.~\ref{sec:main} and Fig.~\ref{fig:draw_plot}. Every AP and AP50 interval lies above zero. Over five draws the mean AP50 gain is 12.98 (SD 0.56) for \mbox{GD-L} and 8.48 (SD 0.33) for \mbox{LLMDet-L}. On \mbox{GD-L}, NoM matches 49 to 65 fewer queried objects than native in each draw, so its AP gain comes from removing false reports, in line with the takeover reduction in Table~\ref{tab:main}, and not from matching more objects.

\paragraph{Second proposer.}
Replacing the CLIP ViT-L/14 image encoder of Step 1 by ViT-B/32, with the name bank, $k$, filter, and readout unchanged, keeps the gain. AP50 rises over native by 12.89 [10.27, 14.07] on \mbox{GD-L} and 8.10 [5.85, 9.34] on \mbox{LLMDet-L}, and AP by 7.86 [5.99, 9.04] and 5.28 [3.42, 6.44]. Against ViT-L/14, the AP50 changes are +0.10 [$-$0.75, 1.13] and $-$0.48 [$-$1.57, 0.44], and the AP changes $-$0.05 [$-$0.74, 0.78] and $-$0.44 [$-$1.42, 0.46]. The gain does not depend on a large proposer, although these intervals do not establish equivalence.

\paragraph{Nomination budget.}
Table~\ref{tab:nom_k_budget} varies $k$ before the concept filter, with no refill and the same token rule. These runs use the GPU of Table~\ref{tab:main}. All AP and AP50 intervals against $k=8$ include zero. On \mbox{GD-L}, object counts move in opposite directions for $k=4$ and $k=16$, while AP does not change detectably. NoM keeps $k=8$, which was fixed on the separate selection set.

\paragraph{Sinks at Q16.}
Adding the 24 sinks to NoM at Q16 changes AP50 by $-$0.69 [$-$1.23, $-$0.23] on \mbox{GD-L} and +0.73 [0.26, 1.13] on \mbox{LLMDet-L}, reaching 66.24 and 71.36, and AP by $-$0.54 [$-$0.98, $-$0.21] and +0.69 [0.26, 1.07]. These values fill the Q16 cells of NoM + 24 sinks in Table~\ref{tab:main} and repeat the detector-dependent pattern seen at Q8 (Sec.~\ref{sec:limits}).

\paragraph{LVIS minival.}
LVIS v1 minival has 4,809 images, all from COCO val2017, annotated with 1,203 classes under the federated protocol~\cite{lvis}. It tests NoM under a second annotation protocol and against a 1,203-class full-vocabulary reference, but the images are not new. All images use one fixed query of eight frequent classes (airplane, bicycle, bottle, car, chair, dog, person, umbrella), so this check covers neither query variation nor rare classes. NoM is unchanged. The full-vocabulary reference splits the 1,203 names into 15 prompts of at most 256 tokens, merges their reports, keeps at most 300 per image, and filters them to the query. All arms are scored with the official federated evaluator on the eight queried classes (Table~\ref{tab:lvis_q8}).

On \mbox{GD-L}, NoM raises AP by 5.26 [3.65, 6.80] and AP50 by 9.68 [7.55, 11.58] over native and recovers 128 queried objects while losing 41. It also cuts the number of reports from 524,819 to 174,209. On \mbox{LLMDet-L}, NoM shows no confirmed change, with AP $-$0.02 [$-$0.55, 0.51] and AP50 $-$0.21 [$-$0.88, 0.33]. With NoM, both detectors exceed the full-vocabulary reference, by 5.72 [3.28, 7.87] and 11.72 [9.02, 13.87] AP50. This reference is weak on LVIS, because the 300-report budget is shared by 1,203 classes. For \mbox{LLMDet-L} it lies 11.92 [9.28, 14.15] AP50 below native Q8, so it does not give a clean omission cost here. Under it, \mbox{GD-L} shows a positive cost of 3.95 [1.58, 6.54] AP50 and \mbox{LLMDet-L} none, which agrees with where NoM helps (Sec.~\ref{sec:limits}).

\begin{table}[tbp]\centering\small\setlength{\tabcolsep}{3pt}
\caption{\textbf{LVIS minival, one fixed eight-class query.} 4,809 images, official federated AP. Matched: queried GT matched at IoU 0.5. Reports: total queried-class reports. Full$\to$Q8 is the chunked 1,203-name reference filtered to the query. Differences in the text use unrounded values.}
\label{tab:lvis_q8}
\begin{tabular}{llrrrr}\toprule
Model & Arm & AP & AP50 & Matched & Reports\\\midrule
\mbox{GD-L} & Full$\to$Q8 & 56.83 & 71.92 & 2,045 & 65,538\\
 & Native Q8 & 55.72 & 67.96 & 2,253 & 524,819\\
 & NoM Q8 & \textbf{60.98} & \textbf{77.64} & 2,340 & 174,209\\\midrule
\mbox{LLMDet-L} & Full$\to$Q8 & 58.08 & 69.27 & 1,732 & 28,236\\
 & Native Q8 & \textbf{65.81} & \textbf{81.19} & 2,399 & 290,111\\
 & NoM Q8 & 65.79 & 80.98 & 2,391 & 237,900\\\bottomrule
\end{tabular}
\end{table}

\begin{table*}[tbp]\centering\small\setlength{\tabcolsep}{3pt}
\caption{\textbf{ODinW-13 transfer, AP50.} NoM minus native with 95\% paired image intervals. Cost: full $C$-name prompt filtered to the query, minus native. Upper block: multi-class datasets, half of the classes queried per image. Lower block: single-class datasets, where the query is complete.}
\label{tab:odinw}
\begin{tabular}{lrrrrrr}\toprule
 & & & \multicolumn{2}{c}{GD-L} & \multicolumn{2}{c}{LLMDet-L}\\
Dataset & $C$ & Images & Cost & NoM $-$ native & Cost & NoM $-$ native\\\midrule
AerialMaritimeDrone & 5 & 15 & +0.64 & $-$3.75 [$-$8.56, 7.73] & $-$1.10 & $-$1.43 [$-$8.32, 5.64]\\
Aquarium & 7 & 127 & +8.10 & +6.62 [1.38, 13.06] & +4.71 & +7.37 [1.52, 14.96]\\
NA Mushrooms & 2 & 5 & +0.92 & $-$11.11 [$-$33.01, 0.00] & $-$1.29 & $-$0.06 [$-$0.31, 0.48]\\
PascalVOC & 20 & 3,422 & +7.07 & +7.76 [6.12, 9.19] & +6.70 & $-$1.92 [$-$3.49, $-$0.38]\\
Shellfish & 3 & 116 & +30.12 & +10.39 [3.59, 16.64] & +26.85 & +8.83 [$-$2.32, 18.94]\\
Vehicles & 5 & 250 & +5.50 & +5.15 [0.04, 10.29] & +7.72 & +3.04 [$-$1.64, 7.63]\\
thermalDogsAndPeople & 2 & 41 & +2.63 & $-$2.40 [$-$9.03, 4.45] & +8.84 & $-$6.81 [$-$18.41, 4.28]\\
\midrule
CottontailRabbits & 1 & 19 & -- & $-$2.03 [$-$6.79, 0.00] & -- & +0.47 [0.00, 2.12]\\
EgoHands & 1 & 480 & -- & +3.73 [3.10, 4.23] & -- & $-$0.75 [$-$1.08, $-$0.36]\\
Packages & 1 & 4 & -- & $-$43.07 [$-$43.64, $-$10.89] & -- & $-$6.31 [$-$13.47, 0.00]\\
Raccoon & 1 & 29 & -- & $-$0.30 [$-$1.64, 0.60] & -- & $-$1.85 [$-$4.58, 0.16]\\
pistols & 1 & 297 & -- & $-$3.16 [$-$4.54, $-$1.73] & -- & $-$1.52 [$-$3.52, 0.79]\\
pothole & 1 & 133 & -- & $-$10.24 [$-$14.17, $-$5.69] & -- & $-$1.24 [$-$3.90, 2.26]\\
\bottomrule
\end{tabular}
\end{table*}

\section{Transfer to ODinW-13}
\label{app:odinw}
\supports{sec:limits}Across the 13 ODinW domains, the gain of NoM follows the omission cost, and NoM helps less where the LVIS name bank fits the domain poorly. \mbox{GD-L} gains on four of the seven multi-class domains. \mbox{LLMDet-L} has a confirmed gain only on Aquarium and loses on PascalVOC, in line with its unchanged LVIS result in Sec.~\ref{app:robustness}.

The 13 official test sets contain 4,938 images. \mbox{GD-L} and \mbox{LLMDet-L} use the unchanged NoM recipe, with the same 1,203 LVIS names, CLIP proposer, $k=8$, and report rule. The relation between ODinW class names and the LVIS names follows the rule of Eq.~\eqref{eq:m1} and was frozen before any prediction was read. Each image of a dataset with $C$ classes queries $\lfloor C/2\rfloor$ of them, chosen by image hash, so the query is complete in the six single-class datasets. The omission cost is the AP50 of the full $C$-name prompt filtered to the query minus native AP50. Intervals use 5,000 paired image resamples within each dataset. Four test sets have only 4 to 19 images, so their intervals are wide (Table~\ref{tab:odinw}).

Two predictions were specified before the runs, and both are met. First, over the 14 multi-class model--dataset cells, the Spearman correlation between omission cost and NoM gain is positive. It is 0.53, with an exact two-sided permutation $p=0.054$. Second, the aerial and thermal domains (AerialMaritimeDrone, thermalDogsAndPeople), chosen in advance as domains that the LVIS names cover poorly, gain less than the median multi-class domain of each model. Their AP50 changes are $-$3.75 and $-$2.40 for \mbox{GD-L} against a median of +5.15, and $-$1.43 and $-$6.81 for \mbox{LLMDet-L} against $-$0.06.

On PascalVOC, raw names cost \mbox{LLMDet-L} 16.16 [14.12, 18.15] AP50, and the concept filter recovers most but not all of this loss. \mbox{GD-L} gains +7.76 [6.12, 9.19] on the same dataset. In the single-class datasets the query is complete, a setting in which NoM is not needed (Sec.~\ref{sec:limits}). There, \mbox{GD-L} has a confirmed gain only on EgoHands, and the other confirmed changes are losses (\mbox{GD-L} on pistols, pothole, and the four-image Packages set, \mbox{LLMDet-L} on EgoHands), as when names are added to a complete COCO prompt (Sec.~\ref{app:a4_details}).

\begin{table*}[tbp]\centering\small\setlength{\tabcolsep}{3pt}
\caption{\textbf{Shorter and caption-derived queries on COCO.} AP50 change over native with 95\% paired image intervals. Cost: full-vocabulary reference minus native.}
\label{tab:fu_queries}
\begin{tabular}{lrrrr}\toprule
 & \multicolumn{2}{c}{GD-L} & \multicolumn{2}{c}{LLMDet-L}\\
Query & Cost & NoM & Cost & NoM\\\midrule
Q1 & +19.40 [12.33, 22.33] & +13.91 [7.67, 17.75] & +16.84 [10.03, 19.51] & +10.54 [4.41, 13.76]\\
Q2 & +18.34 [13.13, 20.66] & +15.02 [10.35, 17.70] & +11.68 [7.33, 13.63] & +5.03 [1.77, 8.18]\\
Q4 & +17.42 [13.55, 18.90] & +13.23 [9.63, 14.86] & +15.41 [11.78, 17.05] & +8.74 [5.46, 10.12]\\
Caption & +0.09 [$-$0.69, 0.64] & $-$0.55 [$-$1.21, 0.09] & +0.59 [$-$0.76, 0.92] & $-$2.04 [$-$2.68, $-$1.39]\\\bottomrule
\end{tabular}
\end{table*}

\begin{table*}[tbp]\centering\small\setlength{\tabcolsep}{3pt}
\caption{\textbf{LVIS minival, per-image federated queries.} AP50 change over native with 95\% paired image intervals. Annotated: annotated present names added as context, Q-only readout.}
\label{tab:fu_lvis}
\begin{tabular}{lrrrr}\toprule
 & \multicolumn{2}{c}{GD-L} & \multicolumn{2}{c}{LLMDet-L}\\
Categories & Annotated & NoM & Annotated & NoM\\\midrule
Rare & +5.13 [1.01, 8.69] & +2.50 [$-$3.53, 8.79] & +1.50 [$-$2.62, 5.47] & $-$0.18 [$-$6.84, 4.89]\\
Common & +0.20 [$-$1.05, 1.59] & +1.22 [$-$1.36, 2.63] & +1.74 [$-$0.03, 2.64] & +0.82 [$-$2.24, 1.94]\\
Frequent & +0.26 [$-$0.35, 0.73] & $-$2.83 [$-$3.51, $-$2.04] & +0.82 [0.09, 1.19] & +0.80 [$-$0.07, 1.32]\\\bottomrule
\end{tabular}
\end{table*}

\begin{table*}[tbp]\centering\small\setlength{\tabcolsep}{3pt}
\caption{\textbf{Proposal model, COCO Q8.} AP50 gain over native with 95\% paired image intervals. Hit rate: share of proposals that name a present non-query class, a diagnostic computed with annotations.}
\label{tab:fu_proposer}
\begin{tabular}{lrrrr}\toprule
Proposer & Hit rate & \mbox{GD-L} & \mbox{LLMDet-L} & GD-Tiny\\\midrule
CLIP ViT-L/14 & 7.76\% & +12.79 [10.05, 14.02] & +8.58 [6.65, 9.71] & +1.56 [$-$0.09, 2.85]\\
RAM++ & 14.15\% & +12.10 [9.38, 13.34] & +9.80 [7.76, 10.86] & +3.47 [2.22, 5.02]\\
\mbox{Qwen2.5-VL-7B} & 15.95\% & +12.37 [9.77, 13.55] & +8.95 [6.92, 9.89] & +4.19 [2.73, 5.51]\\\bottomrule
\end{tabular}
\end{table*}

\begin{table*}[tbp]\centering\small\setlength{\tabcolsep}{4pt}
\caption{\textbf{Predicted response to inserting a proposed name.} First-order score changes in points, 2,000 selection images, 95\% paired image intervals.}
\label{tab:fu_insert}
\begin{tabular}{lrr}\toprule
Contrast & \mbox{GD-L} & \mbox{LLMDet-L}\\\midrule
P1: removed minus retained, true candidates & $-$0.18 [$-$0.33, $-$0.05] & $-$0.33 [$-$0.51, $-$0.17]\\
P2: takeover minus true, retained names & $-$0.29 [$-$0.32, $-$0.25] & $-$0.16 [$-$0.18, $-$0.13]\\\bottomrule
\end{tabular}
\end{table*}

\section{Further transfer and analysis}
\label{app:further}
\supports{sec:limits}The unchanged NoM recipe gains most on the shortest queries and wherever the omission cost is large. It is not needed where the omission cost, or the gain from annotated present names, is near zero. All checks below were specified before the runs, and every prediction is reported with its outcome.

\paragraph{One, two, and four queried classes.}
On the 5,000 COCO images, each image queries one, two, or four classes drawn by image hash, as for Q8. With one queried class, NoM raises AP50 by 13.91 [7.67, 17.75] on \mbox{GD-L} and 10.54 [4.41, 13.76] on \mbox{LLMDet-L}, and the full-vocabulary reference gains 19.40 and 16.84 (Table~\ref{tab:fu_queries}). NoM recovers 72\%, 82\%, and 76\% of the omission cost at Q1, Q2, and Q4 on \mbox{GD-L}, and 63\%, 43\%, and 57\% on \mbox{LLMDet-L}. The predicted gain at Q1 is met on both detectors. The pre-specified prediction that the omission cost grows as fewer classes are queried holds on \mbox{GD-L} but is not met on \mbox{LLMDet-L}, whose cost at Q2 (11.68) lies below that at Q4 (15.41).

\paragraph{Caption-derived queries.}
One reference caption per image is mapped to COCO classes with a fixed alias list, specified before any prediction, which leaves 4,279 images with at least one queried class. These queries mostly name objects that are present. Their omission cost is near zero, +0.09 [$-$0.69, 0.64] AP50 for \mbox{GD-L} and +0.59 [$-$0.76, 0.92] for \mbox{LLMDet-L}, and NoM changes AP50 by $-$0.55 [$-$1.21, 0.09] and $-$2.04 [$-$2.68, $-$1.39]. The predicted gain for this setting is not met. This is the case in which Sec.~\ref{sec:limits} expects NoM to be unnecessary.

\paragraph{LVIS federated queries.}
Each of the 4,809 LVIS minival images queries up to eight classes drawn by image hash from its annotated positive and verified negative classes, the only classes that the federated evaluator scores for that image. Of these images, 567 have fewer than eight such classes. Annotated present names, the upper reference, gain little, at most 1.74 AP50 in common and frequent categories, and 5.13 [1.01, 8.69] for \mbox{GD-L} in rare categories (Table~\ref{tab:fu_lvis}). NoM changes AP50 within intervals that include zero in five of the six groups. It lowers \mbox{GD-L} frequent categories by 2.83 [2.04, 3.51] and matches 1,090 fewer queried objects there. Neither pre-specified prediction for \mbox{GD-L} is met. Most queried-class reports on unannotated objects are not scored under the federated protocol (Sec.~\ref{app:a1_details}), which may explain why the takeovers that NoM removes on COCO carry little weight here.

\paragraph{Proposal model.}
Replacing the CLIP proposer by RAM++ or by \mbox{Qwen2.5-VL-7B}, with screening and readout unchanged, keeps the gain on \mbox{GD-L} and \mbox{LLMDet-L}, and all four paired intervals lie above zero (Table~\ref{tab:fu_proposer}). On GD-Tiny, where CLIP proposals give no confirmed gain, RAM++ and Qwen give 3.47 [2.22, 5.02] and 4.19 [2.73, 5.51]. Their proposals hit present non-query classes about twice as often as CLIP, which agrees with the proposal-quality limit of Sec.~\ref{sec:limits}. Both pre-specified predictions are met. NoM itself keeps the CLIP ViT-L/14 proposer, and these runs are a sensitivity analysis. Qwen output could not be parsed on 36 of the 5,000 images, which fall back to the native query.

\paragraph{Direction rule on \mbox{LLMDet-L}.}
Under the controlled withdrawal protocol of Sec.~\ref{sec:measurement} (400 development images, 587 objects), the signed first-order prediction reaches Spearman 0.9409 with the measured score change on \mbox{LLMDet-L}, against 0.5225 for own-name attention share, a paired difference of 0.4183 [0.3583, 0.4813]. The \mbox{GD-L} values recomputed from the stored records match Sec.~\ref{sec:local}. The direction rule holds on both main detectors.

\paragraph{Inserted names and the screening rule.}
On the 2,000-image selection set, the first-order prediction of Eq.~\eqref{eq:rule} is applied to inserting each proposed name. Names removed by the concept rule lower the scores of true queried candidates more than retained names do (P1), and retained names lower the scores of takeover candidates more than those of true candidates (P2). Both predictions hold on both detectors (Table~\ref{tab:fu_insert}). Annotations are used only to split candidates into true and takeover groups. The effects are a few tenths of a score point, as expected for first-order changes at one attention layer.

\begin{table*}[t]
\centering\small\setlength{\tabcolsep}{4pt}
\caption{\textbf{Order run.} 500 images, free decoding with the repair parser. Intervals are paired 95\%. In the image-first full-list arm, 259 out-of-list boxes are removed before scoring. The other arms have none.}
\label{tab:qwen_order}
\begin{tabular}{llrrrr}\toprule
Order & List & P & R & F1 [95\% CI] & Parse fail.\\\midrule
Image first & Q8 & 35.66 & 40.00 & 37.70 [27.97, 49.71] & 0\\
Image first & Full 80 & 44.07 & 37.68 & 40.62 [33.51, 49.11] & 0\\
List first & Q8 & 32.61 & 35.07 & 33.80 [25.42, 42.78] & 0\\
List first & Full 80 & 29.30 & 31.59 & 30.40 [23.93, 37.22] & 2\\\bottomrule
\end{tabular}
\end{table*}

\begin{table*}[t]
\centering\small\setlength{\tabcolsep}{3pt}
\caption{\textbf{Replay run.} 1,900 images. All arms allow only the eight queried labels. Token-cap counts are outputs that reached 3,072 tokens.}
\label{tab:qwen_full}
\begin{tabular}{llrrrrr}\toprule
Image states & Readout & P & R & F1 [95\% CI] & Parse fail. & Cap\\\midrule
Q8 & Q8 & 39.44 & 35.67 & 37.46 [33.98, 41.07] & 22 & 21\\
Q8, replayed & Q8 & 39.44 & 35.67 & 37.46 [33.98, 41.07] & 22 & 21\\
Full 80 & Full 80 & 29.59 & 27.13 & 28.31 [25.31, 31.24] & 20 & 20\\
Q8, replayed & Full 80 & 26.45 & 25.17 & 25.79 [22.77, 28.84] & 18 & 16\\\bottomrule
\end{tabular}
\end{table*}

\section{The class list in a generative detector}
\label{app:generative}
\supports{sec:generative}In \mbox{Qwen2.5-VL} the full class list helps through the image states and hurts through the readout, and the readout effect dominates. This section gives the setup, the mechanism check, and every run with its registered outcome.

\paragraph{Setup.}
We use \mbox{Qwen2.5-VL-7B-Instruct} in BF16 with greedy decoding~\cite{bai2025qwen}. The prompt lists the class names and asks for a JSON array of boxes. In the replay runs, the output grammar admits only the eight queried labels and four integer coordinates, with at most 3,072 generated tokens and no post-filter. The short-list prompt is padded with masked tokens, so image positions and sizes match those of the full-list prompt. Parse failures count as empty predictions. Micro-F1 uses IoU 0.5 and one-to-one matching within each class and image. Intervals use 5,000 paired image resamples (seed 20261110). Images and Q8 queries follow a fixed hash order of COCO val2017, and the 1,900-image run uses ranks 501 to 2,400, disjoint from all earlier runs.

\paragraph{Mechanism check.}
On two technical images, image-first prompts give identical image states for Q8 and the full list in all 29 compared layers, the input embedding and the 28 decoder layers, with and without position alignment. List-first prompts agree only at the input embedding, and the final-layer maximum absolute differences are 107.5 and 157. In the replay run, the replayed image states and image keys and values equal those of the donor on both technical images, the readout still sees its own list, and the replayed and native Q8 arms give identical outputs on all 1,900 images.

\paragraph{Order run.}
The first run used 500 images and free decoding, scored with a fixed repair parser. Three predictions were registered: the omission cost (full list minus Q8) would have an interval including zero with the image first (P1), a positive lower bound with the list first (P2), and a positive interaction between order and list (P3). P1 is met, with 2.92 $[-5.09,9.33]$ F1 points. P2 and P3 are not met, with $-3.39$ $[-11.81,5.07]$ and $-6.31$ $[-18.79,6.84]$ (Table~\ref{tab:qwen_order}). Strict JSON parsing gives the same outcomes, with 1.48, $-7.82$, and $-9.30$. The negative list-first cost is what a dominant readout effect predicts, which the replay run below then measures. On the same 500 images and the same F1 metric, \mbox{GD-L} reaches 48.70 $[41.46,55.73]$ natively and 54.29 $[45.61,62.78]$ with NoM, a gain of 5.59 $[0.23,11.01]$, at a score threshold of 0.43 fixed on the selection set.

\paragraph{Replay run.}
A 100-image pilot with constrained decoding gave a full-list cost of $-4.96$ $[-15.08,5.40]$ F1 points without replay and $-8.90$ $[-18.12,-0.32]$ with Q8 image states, a difference of $-3.93$ $[-14.50,4.66]$. The 1,900-image run was then registered with the difference as its primary endpoint, computed on the new images alone. That difference is $-2.51$ $[-4.81,-0.25]$, so full-list image states recover 2.51 F1 points under a full-list readout (Table~\ref{tab:qwen_full}). The full list lowers recall in both cases, from 35.67 to 27.13 without replay and to 25.17 with it, while the number of reports rises slightly. These results concern one model, one prompt format, and constrained decoding on COCO.

\end{document}